\documentclass[acmlarge,screen]{acmart}

\AtBeginDocument{%
  \providecommand\BibTeX{{%
    \normalfont B\kern-0.5em{\scshape i\kern-0.25em b}\kern-0.8em\TeX}}}

\setcopyright{rightsretained} 
\acmJournal{IMWUT}
\acmYear{2022} \acmVolume{6} \acmNumber{3} \acmArticle{107} \acmMonth{9} \acmPrice{}\acmDOI{10.1145/3550289}

\usepackage{subfig}
\usepackage{booktabs}
\usepackage{multirow}
\usepackage{tcolorbox}
\usepackage{enumitem}
\usepackage{amsmath}
\usepackage{algorithm}
\usepackage{algpseudocode}
\usepackage{bbm}
\usepackage[normalem]{ulem}

\begin{document}

\newcommand{\parjump}{\vspace{+0.5em}}
\newcommand{\system}{FLAME}
\newcommand{\problem}{MDE}
\newcommand{\dead}{invalid}
\newcommand{\Dead}{Invalid}
\newcommand{\revision}[1]{\textcolor{black}{#1}}
\newcommand{\ak}[1]{\textcolor{red}{\textbf{Akhil: } #1}}
\newcommand{\hy}[1]{\textcolor{cyan}{\textbf{Hyunsung: } #1}}
\newcommand{\todo}[1]{\textcolor{green}{\textbf{TODO: } #1}}

\title{FLAME: Federated Learning Across Multi-device Environments}

\author{Hyunsung Cho}
\authornote{This work was done while the author was an intern at Nokia Bell Labs Cambridge.}
\affiliation{%
  \institution{Carnegie Mellon University}
  \city{Pittsburgh}
  \state{Pennsylvania}
  \country{USA}
}
\email{hyunsung@cs.cmu.edu}
\orcid{0000-0002-4521-2766}

\author{Akhil Mathur}
\affiliation{%
  \institution{Nokia Bell Labs}
  \city{Cambridge}
  \country{UK}}
\email{akhil.mathur@nokia-bell-labs.com}
\orcid{0000-0002-1475-3017}

\author{Fahim Kawsar}
\affiliation{%
  \institution{Nokia Bell Labs}
  \city{Cambridge}
  \country{UK}}
\email{fahim.kawsar@nokia-bell-labs.com}
\orcid{0000-0001-5057-9557}

\renewcommand{\shortauthors}{Cho et al.}

\begin{abstract}
Federated Learning~(FL) enables distributed training of machine learning models while keeping personal data on user devices private. While we witness increasing applications of FL in the area of mobile sensing, such as human activity recognition~(HAR), FL has not been studied in the context of a \textit{multi-device environment}~(MDE), wherein each user owns multiple data-producing devices. With the proliferation of mobile and wearable devices, MDEs are increasingly becoming popular in ubicomp settings, therefore necessitating the study of FL in them. 
FL in MDEs is characterized by being not independent and identically distributed (non-IID) across clients, complicated by the presence of both user and device heterogeneities. Further, ensuring efficient utilization of system resources on FL clients in a MDE remains an important challenge. In this paper, we propose \system{}, a user-centered FL training approach to counter statistical and system heterogeneity in MDEs, and bring consistency in inference performance across devices. 
\system{} features (i) user-centered FL training utilizing the time alignment across devices from the same user; (ii) accuracy- and efficiency-aware device selection; and (iii) model personalization to devices. We also present an FL evaluation testbed with realistic energy drain and network bandwidth profiles, and a novel class-based data partitioning scheme to extend existing HAR datasets to a federated setup. Our experiment results on three multi-device HAR datasets show that \system{} outperforms various baselines by \revision{4.3-25.8}\% higher $F_1$ score, 1.02-2.86$\times$ greater energy efficiency, and up to \revision{2.06}$\times$ speedup in convergence to target accuracy through fair distribution of the FL workload. 
\end{abstract}



\maketitle

\section{Introduction}
\label{sec:introduction}
{Federated Learning (FL)}~\cite{mcmahan2017communication,fedopt} enables collaborative training of machine learning models in networks of remote devices (or clients), while keeping users' personal data on the devices private. At a high level, FL involves repeating three steps: (i) updating the parameters of a shared prediction model locally on each remote client, (ii) sending the local parameter updates to a central server for aggregation, and (iii) receiving the aggregated prediction model back on the remote client for the next round of local updates. While many of the early use-cases of FL were related to natural language processing~\cite{47586}, visual recognition~\cite{he2021fedcv, liu2020fedvision}, and speech recognition~\cite{hard2020training} tasks, we are now also witnessing its applications in the area of human-activity recognition~\cite{doherty2017large}.

Prior FL works have made a common assumption that each remote user owns a \emph{single data-producing device}, which then participates as a client in the federated training process. However, there is nowadays a trend of people simultaneously using \emph{multiple data-producing devices} such as smartphones, smartwatches and smart earbuds to collect data about their physical activities, health, or context. It is even predicted that by 2025, each person will own 9.3 connected devices on average~\cite{deviceby2025}. From a sensing perspective, multi-device environments offer exciting opportunities to develop accurate and generalizable models by leveraging the similarities and differences across devices.

Although there have been prior works on training machine learning models for multi-device environments~\cite{sztyler2017position, min2019closer, jain2022collossl}, they were primarily based on centralized training and required sharing of raw data between devices. Applying federated learning to these settings could be a potential privacy-preserving solution, however to the best of our knowledge no prior works have investigated federated learning in these settings. 

The presence of multiple devices on a user presents the following research challenges and opportunities for federated learning:

\begin{figure}[t]
    \centering
    \vspace{-0.4cm}
    \subfloat[Single-device FL]{\label{smdfl-a}\includegraphics[width=0.4\linewidth]{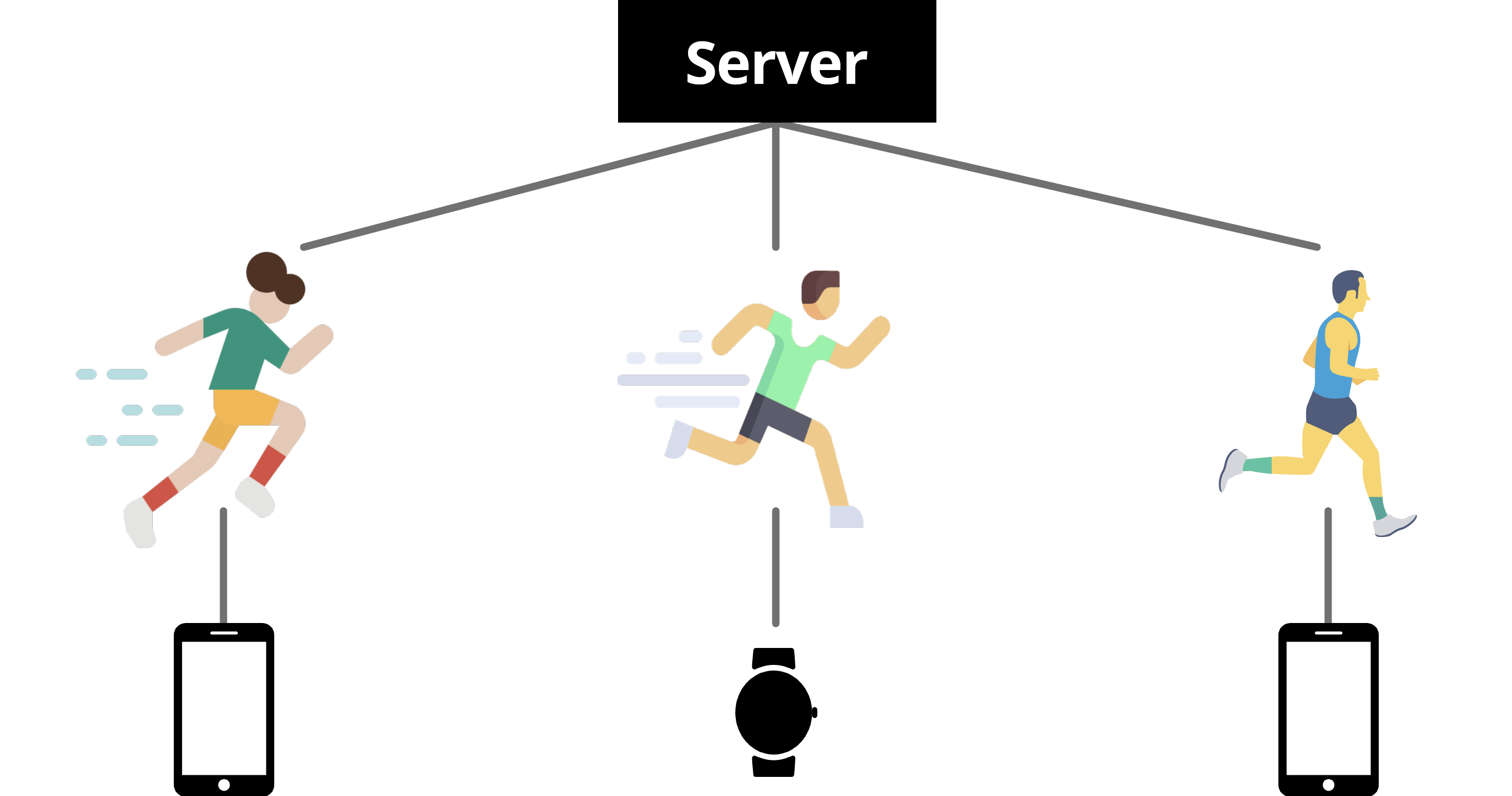}}\hfill
    \subfloat[Multi-device FL]{\label{smdfl-b}\includegraphics[width=0.4\linewidth]{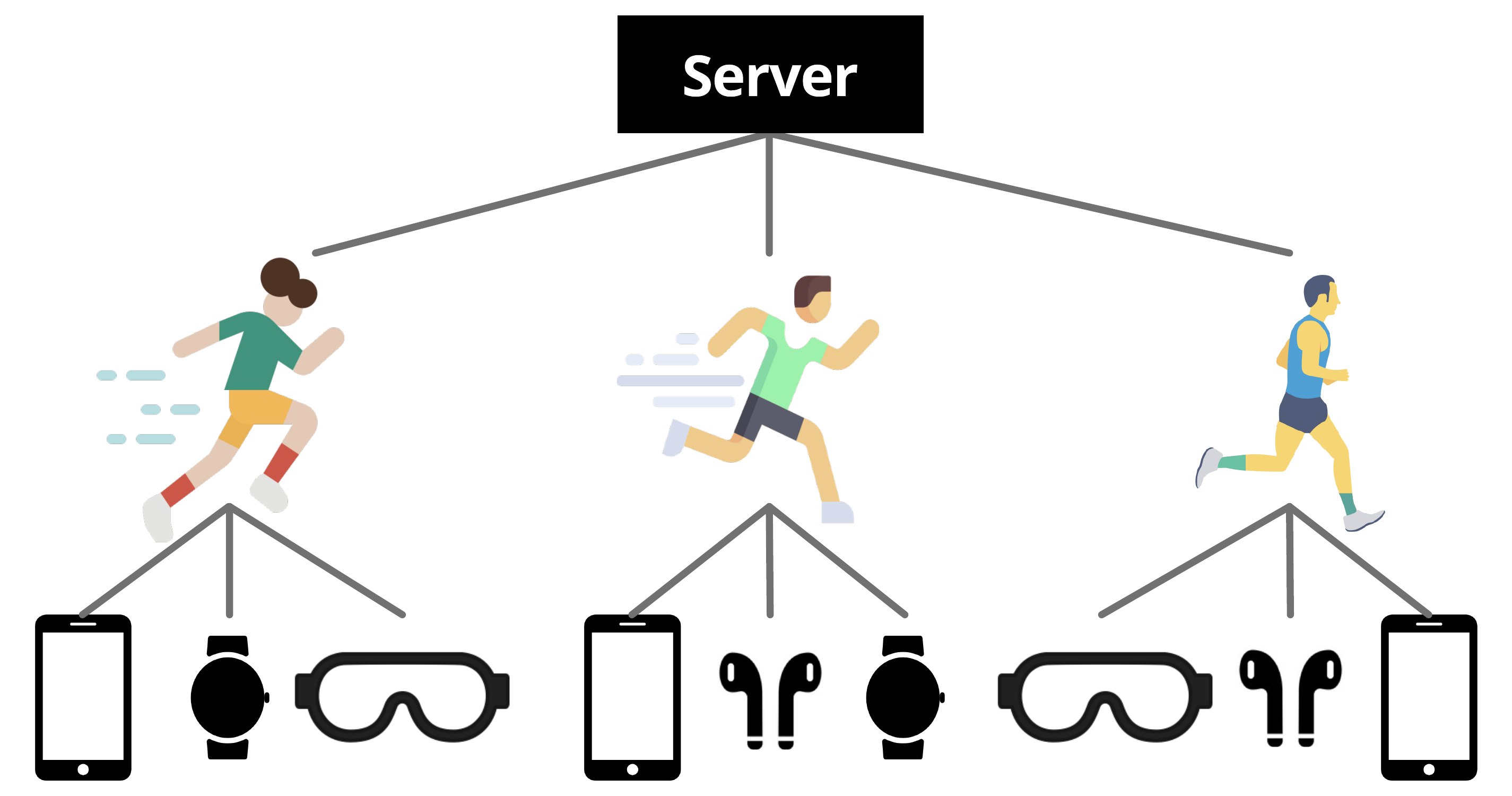}}
    \vspace{-0.2cm}
    \caption{Architectures of single-device FL and multi-device FL. In single-device FL, each user is assumed to have a single data-producing device, whereas in multi-device FL, multiple devices observe the user's activity simultaneously.}
    \vspace{-0.2cm}
    \label{fig:single_multi_fl}
\end{figure}

\begin{itemize}[leftmargin=*]
\setlength\itemsep{0.4em}

\item A multi-device environment can have $N$ users, each owning multiple ($K\geq 1$) data-producing devices that simultaneously collect sensor data (see Figure~\ref{smdfl-b}). Conventional FL systems would consider these $N \cdot K$ devices as independent FL clients; however, this approach ignores the natural affinities in the data collected by the devices of the same user. A promising alternative is to offload the data from all $K$ devices of each user on an edge node (e.g., a smartphone or home router) and treat it as a unified dataset that {represents} the user's behavior. One can then locally training ML models on this dataset, or perform FL considering each of the $N$ consolidated datasets as a separate client. This approach, however, assumes that devices -- possibly from different manufacturers -- will share their raw sensor data with each other. Such data sharing may not happen due to privacy or proprietary reasons, thereby limiting the applicability of this approach. Hence, in this work, we seek to design a federated learning approach which respects the privacy of each device's data and at the same time considers the natural affinity between the devices of the same user.

\item Countering the statistical heterogeneity across users during training is an active area of research in FL~\cite{scaffold, dirchsplit, noniidfl, flgn}. The multi-device problem setting presents a unique challenge that there exists statistical heterogeneity not just across user datasets, but also within a user's local dataset. This `local' statistical heterogeneity is caused by the differences in data distributions of the various devices owned by the user. For example, motion sensors placed at different positions of the body will capture the user's motion differently, thereby making the device datasets heterogeneous. In \S\ref{sec:problem}, we quantify the impact of both these types of heterogeneity in multi-device FL systems and propose a heterogeneity-aware client selection approach to mitigate it. 

\item In addition to the statistical heterogeneity, there also exists system heterogeneity across devices, e.g., variations in computational capabilities, battery power, network communication speeds. These variations are present both across users and across devices of the same user. As system efficiency is a core success metric for a FL system, we need to strike a balance between model accuracy, convergence speed, and resource consumption on devices. In \S\ref{sec:client_selection}, we detail a strategy which combines both statistical and systems utility of each device in a unified metric to drive the client selection in FL.  

\item Due to the high statistical heterogeneity in the multi-device setting, it is likely that a single global model will not generalize to all the devices, and it may even result in uneven accuracies across devices of the same user. We elaborate on this challenge in \S\ref{sec:uneven_accuracy} and present a weight-regularized federated personalization approach to train accurate models tailored to each device.  
\end{itemize}

\parjump{}
\noindent
The main contributions of this paper are as follows:

\parjump{}
\noindent
\textbf{Extending FL to Multi-device Environments.} We present Federated Learning Across Multi-device Environments~(\system{}), a unified solution to solve the aforementioned challenges for FL in multi-device environments. \system{} employs a user-centered FL training approach in combination with a device selection scheme that balances accuracy, convergence time, and energy efficiency of FL. \system{} further utilizes model personalization to counter statistical heterogeneity and inconsistency in inference performance across devices.

\parjump{}
\noindent
\textbf{A Realistic Testbed for FL.} Prior works on federated training of HAR models (e.g., ~\cite{yu2021fedhar}) have assumed that each FL client owns a large amount of labeled data, in the order of 3000-5000 seconds. Instead, we setup a realistic FL testbed with a large number of clients, each holding a small amount of labeled data. This is achieved through a novel class-based partitioning scheme that divides existing HAR datasets over a large number of users, and makes them suitable for realistic federated evaluations. In addition, our testbed includes latency and energy consumption profiles for nine embedded-scale hardware, which allows for obtaining a realistic estimate of the resource consumption of federated HAR algorithms.  
We plan to open-source the source code of our partitioning algorithm as well as the federated HAR datasets for reproducibility.

\parjump{}
\noindent
\textbf{Extensive Empirical Analysis.} We compare \system{} against various FL baselines in terms of inference performance, training time, and energy consumption on three multi-device HAR datasets. These datasets contain inertial sensor data for human activities ranging from locomotion tasks to activities of daily living. For a deeper analysis of \system{}, we also present sensitivity and ablation studies on it. Our results highlight the superior inference performance, energy-efficiency, and convergence rate of \system{} as compared to the baselines. 

\section{Background} \label{sec:primer}

In this section, we provide a primer on Federated Learning (FL) and explain the factors that influence the performance of FL systems. 

\subsection{Key Elements of a Federated Learning System}
\label{subsec.primer}
A federated learning system consists of a central server and $N$ remote clients, often containing labeled data samples. The central server randomly initializes a deep neural network model $M^0$ and sends it to a subset of the clients $C \subseteq N$ for local training. Each client $c \in C$ updates the parameters of the model $M^0$ independently through supervised learning, by optimizing a loss function such as categorical cross-entropy on its local dataset. In practice, the local training is done by iterating over the labeled dataset $E$ times, where $E$ denotes the number of \emph{local epochs}.  After the local training finishes, we obtain a trained model $M^0_c$ on each client $c$. These local models are sent to the central server, where the model parameters are averaged using federating averaging algorithms such as FedAvg~\cite{BrendanMcMahan2017} to obtain the new global model $M^1$. This entire process of local training and federated averaging is repeated for $R$ rounds to obtain the final global model $M^R$.

There are three key evaluation metrics for a FL system: {firstly}, we would like the model $M^R$ to have high \textbf{test accuracy} on each client. Secondly, we would like the training to be \textbf{time-efficient}, both in terms of the number of rounds taken to convergence and the overall wall-clock time of training. Finally, we would like the training to be \textbf{energy-efficient}, in that it should minimize the battery drain due to local training on remote clients. 

\parjump{}
\noindent
Below we describe two important factors that influence the performance of an FL system. 

\begin{itemize}[leftmargin=*]

\item \textbf{IID-ness of client datasets.} The \emph{test accuracy} of a model learned using FL is adversely impacted if the datasets on the remote clients are not independent and identically distributed (IID)~\cite{noniidfl, zhu2021federated}. The fundamental reason for this performance degradation is that when the client datasets are non-IID or heterogeneous, the local models trained on the clients may diverge, despite having the same initial parameters. This parameter divergence in local models, in turn, makes the global model obtained after federated averaging sub-optimal, thereby worsening its test accuracy.

\item \textbf{Clients selected in each round.} In each round of FL, the server samples $C \subseteq N$ clients for local training. The choice of sampled clients can impact FL performance in three ways: (a) the data heterogeneity across sampled clients can influence the global model's \emph{accuracy}, (b) the communication bandwidth and computational resources of the sampled clients have a direct influence on how fast the local model is trained and communicated back to the global server. This in turn will impact the \emph{training time} of FL, and (c) excessive local training on a client can lead to severe \emph{battery drain}, thus leading to {failure cases in selected clients and} user experience issues such as devices running out of battery power. 

\parjump{}
\noindent
Addressing the challenges of non-IID-ness or statistical heterogeneity~\cite{lisurvey, dirchsplit, noniidfl} and achieving higher system efficiency through better client selection~\cite{fedprox, oort} are active topics of research in FL.

\end{itemize}

\section{Federated Learning in Multi-Device Environments} \label{sec:problem}
Building on the primer presented in \S\ref{subsec.primer}, we now contextualize FL in a multi-device environment and aim to highlight the challenges and research opportunities for FL in this scenario. First, we define and state our assumptions about a multi-device environment.  

\parjump{}
\noindent
\textbf{Multi-Device Environment (MDE):} In this setup, there exist \emph{multiple} sensor devices that capture a physical phenomenon, e.g., a user's locomotion activity, {simultaneously}. In the context of human-activity recognition, an example of MDE is when a person wears multiple inertial sensing devices on their body~\cite{hasan2019comprehensive} as shown in Figure~\ref{smdfl-b}. These multiple devices observe the user's activity or context \emph{simultaneously} and record sensor data in a \emph{time-aligned manner}~\cite{sztyler2017position, jain2022collossl}. In contrast, conventional federated learning setups assume that each user (or client) has a single data-producing device as shown in Figure~\ref{smdfl-a}.

\begin{figure}[t]
    \centering
    \includegraphics[width=\textwidth]{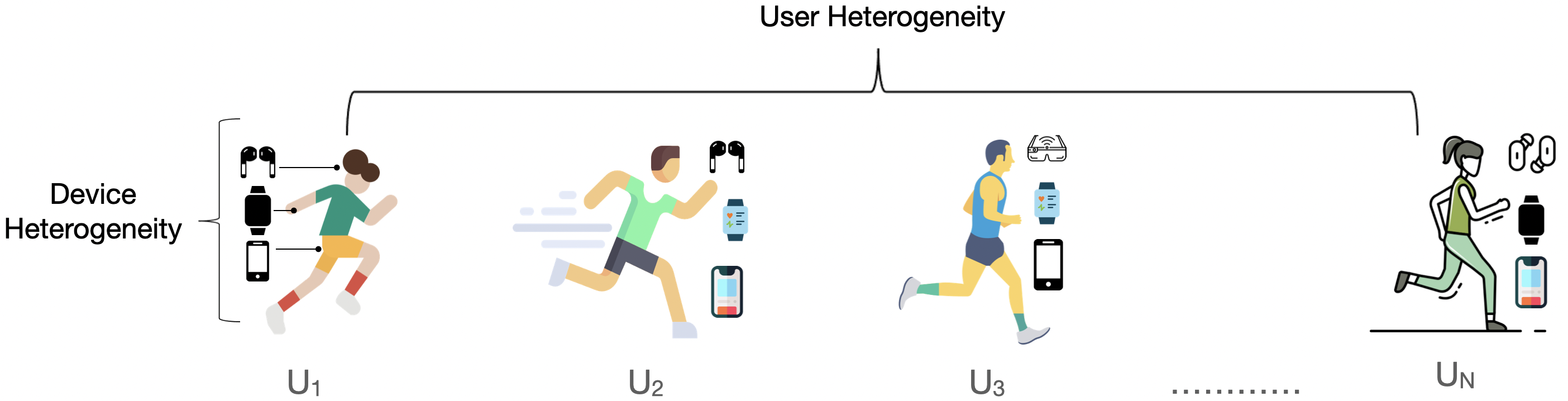}
    \vspace{-0.3cm}
    \caption{Illustration of user and device heterogeneities in a multi-device FL system. $U_1 \cdots U_N$ are different users, each wearing multiple heterogeneous data-producing devices.}
    \vspace{-0.3cm}
    \label{fig:hetero}
\end{figure}

\parjump{}
\noindent
Below we present three key challenges for federated learning in the \problem{} setup.

\subsection{Device and User Heterogeneity Lead to Higher Non-IID-ness} The \problem{} setting presents a challenging case of non-IID-ness in client datasets because of the presence of two types of data heterogeneities as illustrated in Figure~\ref{fig:hetero}: 
\begin{itemize}[leftmargin=*]
\setlength\itemsep{0.5em}
    \item  \emph{User heterogeneity} occurs due to differences in personal characteristics of the users, such as different running styles and gait variations~\cite{sztyler2017online, wijekoon2020knowledge}. As federated learning, by definition, aims to leverage the data from multiple users to learn a prediction model, this form of heterogeneity is expected across FL clients. 
    
    \item \emph{Device heterogeneity}. In addition to differences across users, the \problem{} setup also exhibits heterogeneity due to the multiple devices owned by the users. This heterogeneity comes from the differences in hardware and software components of the multiple devices worn by the user as well as their positions on the user's body. For example, devices with inertial sensors can be placed on the wrist (smartwatch), in the ear (smart earbud), or inside a trouser pocket (smartphone). Prior works have shown that device heterogeneity can lead to distribution shifts in HAR data~\cite{stisen2015smart, chang2020systematic}. Critically, \emph{device heterogeneity} can be present even within a user's own data when a user wears multiple sensor-enabled devices while performing an activity. 
\end{itemize}

\noindent
To illustrate the effect of these heterogeneities on the non-IID-ness of a FL setup, we present a quantification experiment on the RealWorld HAR dataset~\cite{sztyler2016onbody}. This dataset consists of physical activity data collected from 3-axis accelerometer and 3-axis gyroscope sensors by $N$(=15) users, while they are wearing $K$(=7) IMU-enabled devices placed at thigh, waist, chest, head, shin, upperarm, and forearm. Specifically, we compare two scenarios: (a) when each of the $K$ devices owned by the $N$ users are considered as separate FL clients, and (b) the ideal case when each of the $N$ users only have a single device (such as a wrist-mounted IMU or a head-mounted IMU). While (b) is a conventional FL setup with users owning a single data-producing device, the scenario (a) exemplifies the \problem{} setup with multiple users owning multiple devices.

\begin{figure}[t]
    \begin{minipage}{0.35\textwidth}
        \centering
        \includegraphics[width=0.8\linewidth]{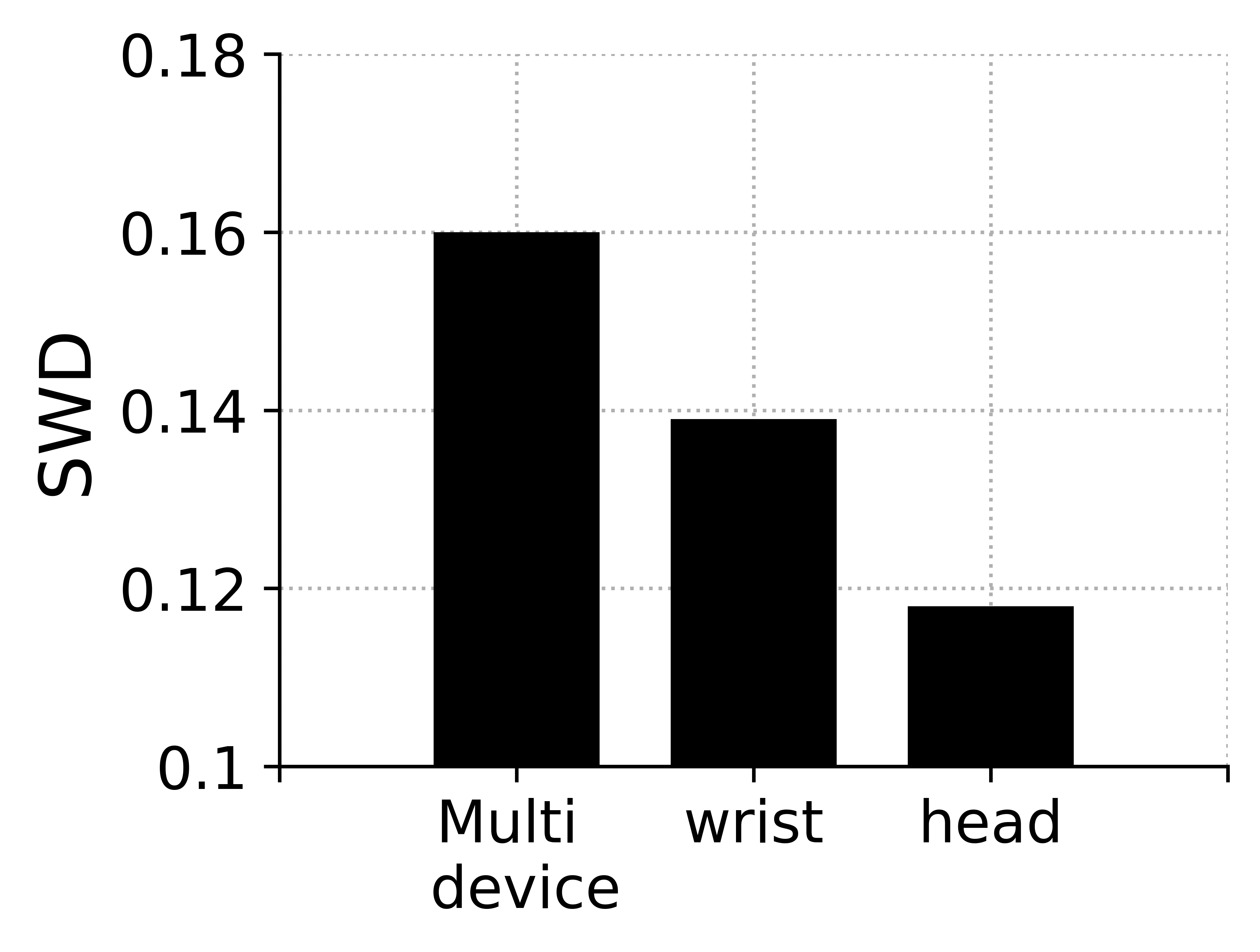}
    \vspace{-0.4cm}
    \captionof{figure}{Comparison of mean Sliced Wasserstein Distance (SWD) in a multi-device FL setup with two single-device FL setups. A higher mean SWD indicates higher non-IID-ness of the FL setup. In the multi-device setup, each user has multiple devices that act as FL clients. The single-device setup exemplifies a conventional FL setup wherein each user has just one device (i.e., wrist-worn IMU or head-worn IMU).}
        \label{fig:swd}
    \end{minipage} \hfill
    \begin{minipage}{0.6\textwidth}
        \centering
        \vspace{-0.4cm}
    \subfloat[]{\label{}\includegraphics[width=0.45\linewidth]{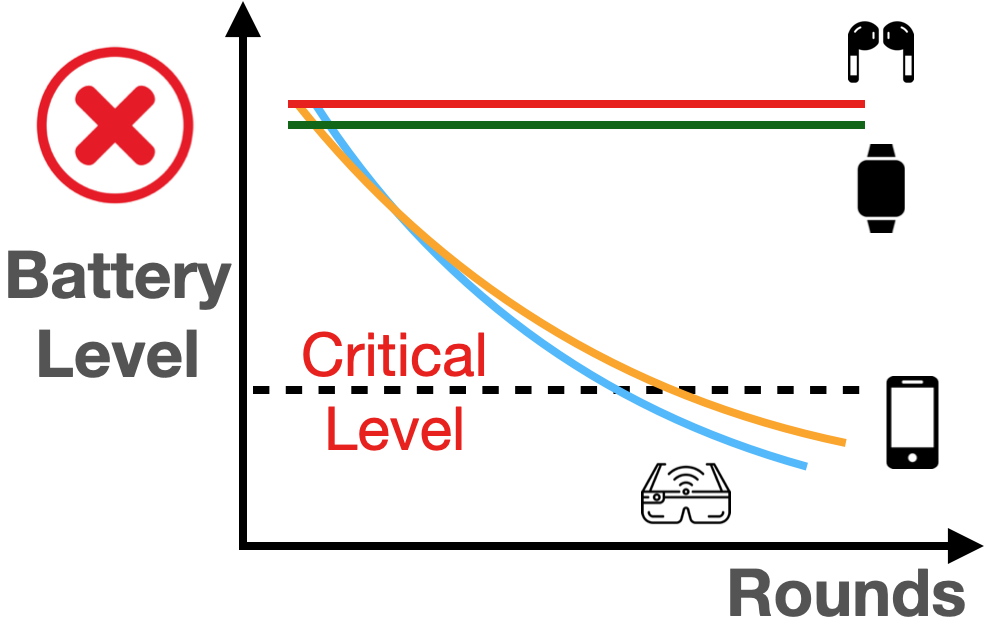}}
    \subfloat[]{\label{}\includegraphics[width=0.45\linewidth]{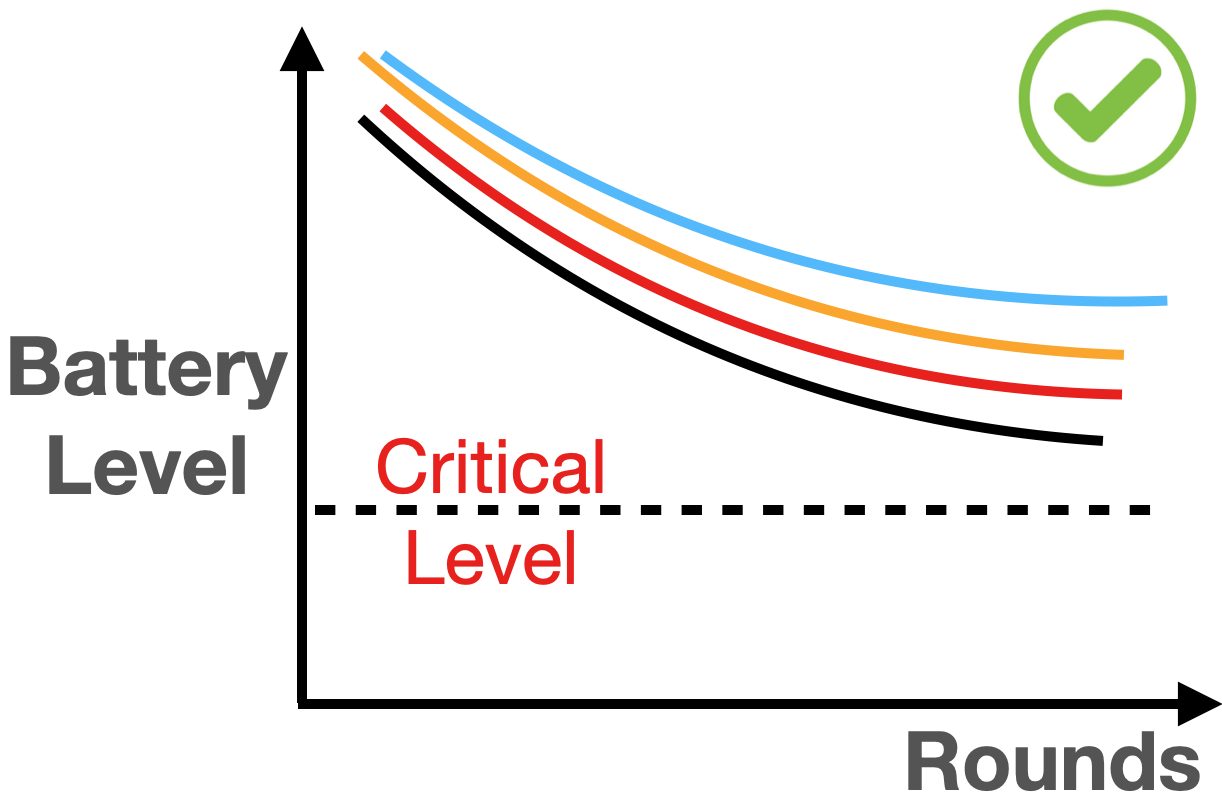}} 
    \vspace{-0.4cm}
    \captionof{figure}{A conceptual illustration to depict the potential impact on local devices if client selection does not balance energy and accuracy. We show the state of four devices owned by a user over the FL training process. (a) Only two of the four devices are ever selected for training as they result in higher model accuracy. This leads to a situation that both these devices get over-sampled, resulting in a significant energy drain due to local training and their battery levels go below a critical level required for other device operations. (b) A better approach which balances the energy consumption of each device during the training process and ensures that none of the devices go below the critical level.}
    \label{fig:conceptual_energy}
    \end{minipage}%
\end{figure}

We use Sliced Wasserstein Distance (SWD)~\cite{bonneel2015sliced, rabin2011wasserstein} as a metric for this quantification. SWD is a metric used to compute the distance between two multi-dimensional data distributions; a higher SWD implies a larger heterogeneity between the distributions. We compute the pairwise SWD for every pair of FL clients\footnote{The SWD is computed over the 6-dimensional accelerometer and gyroscope datasets on each client. We segment the data in windows of 3 seconds based on prior work~\cite{chang2020systematic, liono2016optimal}; this yields a $ X_i \times 6$ dimension dataset on each client, where $X_i$ denotes the number of windows on client $i$.} and report their mean. A higher mean SWD of the entire setup would indicate a higher degree of non-IID-ness across clients. Figure~\ref{fig:swd} illustrates our findings. We observe that the multi-device setup has a mean SWD which is 33\% and 14.2\% higher than the SWDs of single device setups with head- and wrist-mounted IMUs, respectively. This finding confirms that the presence of both user and device heterogeneities in the \problem{} setup leads to a higher degree of non-IID-ness across clients. Finally, we trained FL models in each of these settings to evaluate the impact of non-IID-ness on model accuracy. Our results show that a model trained in the multi-device setting has an accuracy of 54.2\% and 52.6\% on the test sets of head and wrist respectively. In comparison, a model trained in the single-device setups of head and wrist obtain accuracies of 72.4\% and 67.3\% respectively on their test datasets. This finding confirms that the presence of both user and device heterogeneities in the multi-device setup lead to degradation in the performance of FL algorithms.

\subsection{Balancing Accuracy, Energy and Convergence Time} In addition to model accuracy, convergence time and energy consumption are two important evaluation metrics for FL systems. The \emph{convergence time} of FL depends on the computational capabilities of each device and the communication bandwidth between the device and the server, e.g., if a device has a slow processor or a low network bandwidth, it will take more time to train the model and communicate its parameters to the server, which in turn would increase the FL convergence time. Similarly, the \emph{energy consumption} of FL is the sum of energy consumed on each of the client devices during training. Each client's energy cost depends on the power efficiency of its processor (e.g., CPU, GPU) as well as the time taken to complete the local training over all FL rounds. During each round of federated training, we need to ensure that clients are selected in a way that achieves a good balance between model accuracy, convergence time and energy efficiency. Instead, if we optimize for only one of these metrics (e.g., accuracy) during client selection, we could run into undesirable cases as depicted in Figure~\ref{fig:conceptual_energy}. For example, only some of a user's devices would ever get selected as they boost model accuracy. This over-sampling, however, drives quick, concentrated energy consumption in the selected devices below a critical level for operations.

\begin{figure}[t]
\centering
\subfloat{\label{}\includegraphics[width=0.48\linewidth]{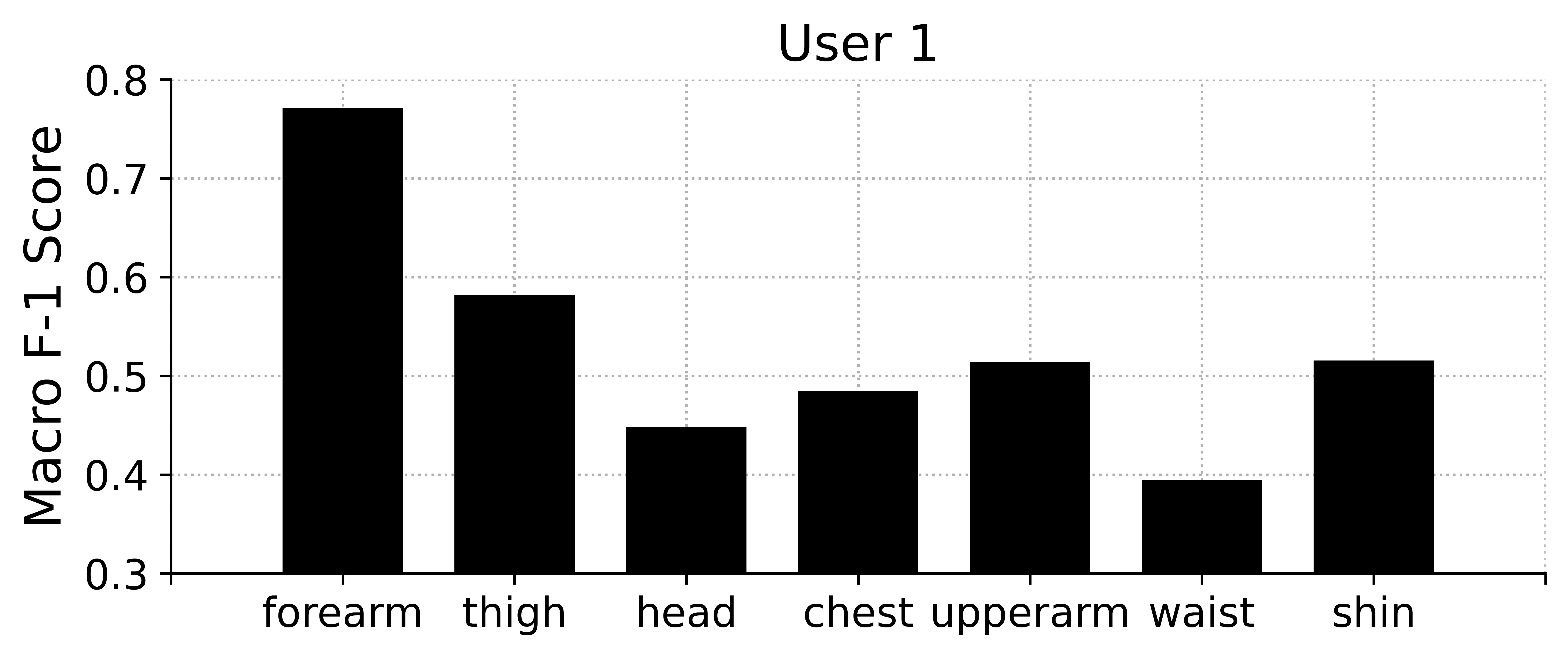}}
\subfloat{\label{}\includegraphics[width=0.48\linewidth]{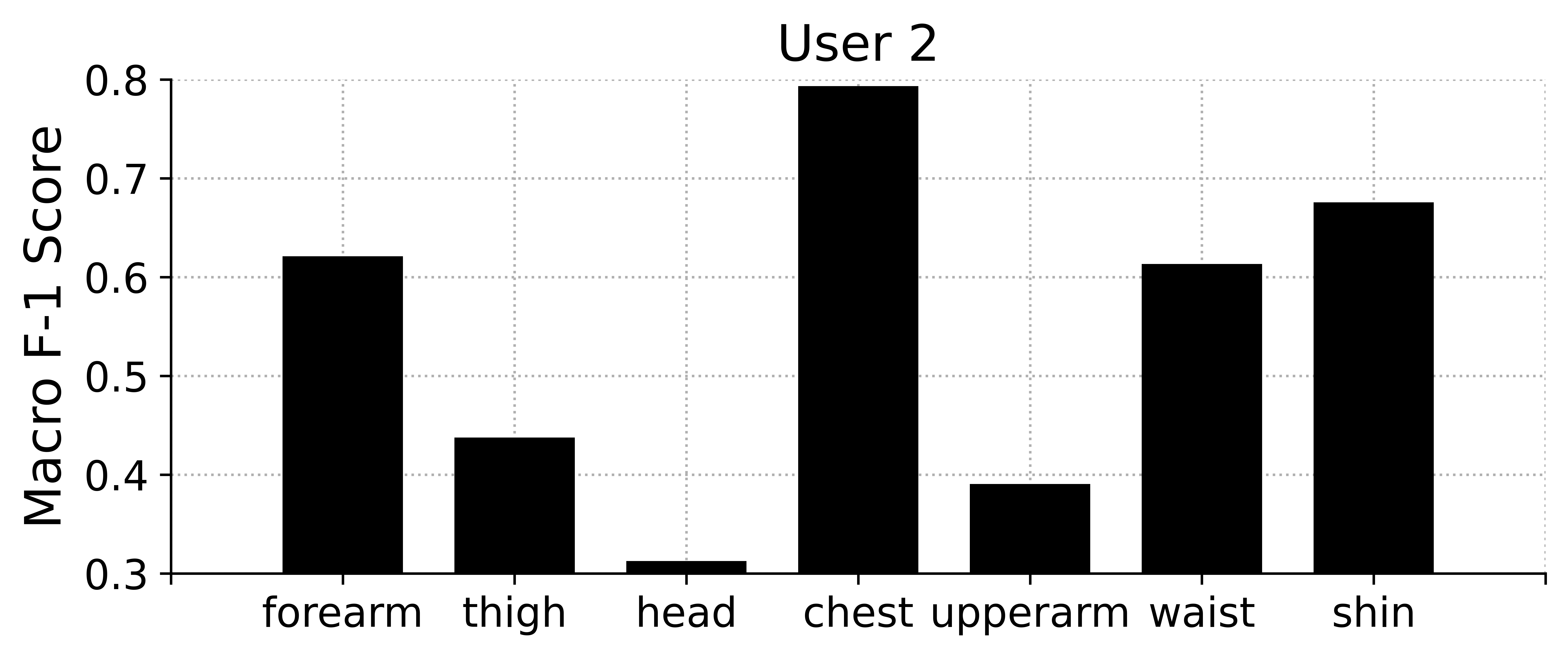}} \par 
\caption{Macro $F_1$ scores obtained using the global model trained for 100 FL rounds. We observe a high variance in model performance across the devices owned by the same user. See \S\ref{sec:hparams} for more details about the experiment setup.}
\label{fig:fedavg_user_variations}
\end{figure}

\subsection{Uneven Accuracy Across Devices and Users} \label{sec:uneven_accuracy}
The output of FL is typically a single global model which is obtained by averaging the local models from all the participating clients. However, in a \problem{} setup with various types of statistical heterogeneity present in the data, the global model may not be optimal for each participating user and device. We illustrate this challenge in Figure~\ref{fig:fedavg_user_variations} where we plot the test accuracies obtained using the global model on devices owned by two randomly sampled users from the RealWorld HAR dataset~\cite{sztyler2016onbody}. The global model in this case is trained for 100 rounds of FL on the RealWorld dataset. We observe a high variance in the macro $F_1$ score obtained for different devices owned by the same user. For example, the chest-worn IMU device of user 2 achieves a test $F_1$ score of 0.8 while the head-worn IMU device only has a 0.31 $F_1$ score using the global FL model. Clearly, such variations in prediction outcomes across devices is not ideal for user experience as they may cause confusion for the user on which outcome to trust.

\vspace{0.2cm}
\begin{tcolorbox}
\textbf{Takeaways:} The \problem{} setup amplifies the challenges for FL by introducing higher non-IID-ness across clients; necessitating a careful balance between accuracy, convergence time and energy consumption; and causing significant variance in performance of different devices of the same user.  
\end{tcolorbox}

\section{\system{}: Federated Learning Across Multi-device Environments} \label{sec:solution}

\begin{figure}[t]
    \centering
    \includegraphics[width=0.9\linewidth]{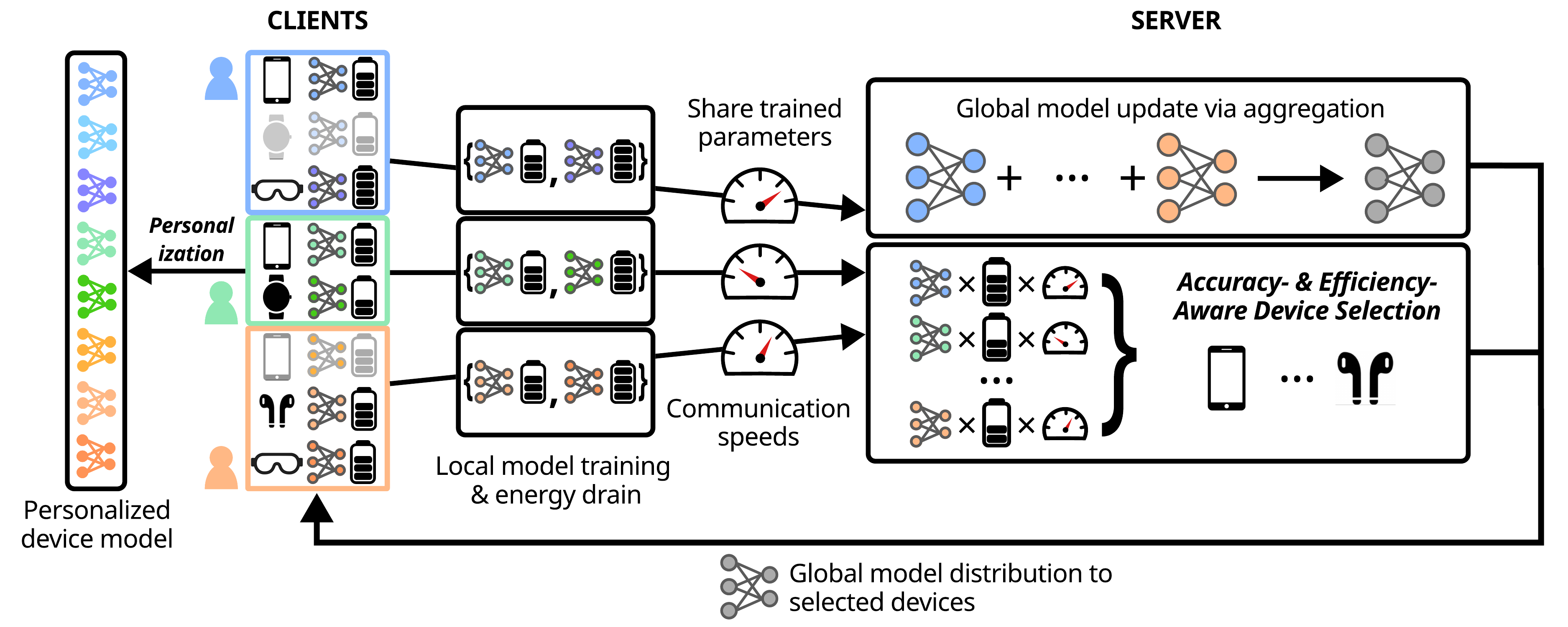}
    \vspace{-0.2em}
    \caption{System architecture and workflow of \system{}. The server distributes the global model to selected client devices. Clients perform training on their local dataset and upload model updates along with their unified utility metric (see \S\ref{sec:client_selection}) to the server. After the upload, clients personalize their device models using the up-to-date global model weights. Meanwhile, the server updates the global model by aggregating local updates from selected clients. It also uses the utility metrics reported by the client to perform accuracy- and efficiency-aware device selection and samples clients for the next round.}
    \label{fig:overview}
\end{figure}

\system{} is our end-to-end FL solution comprising of a novel client selection scheme that minimizes data heterogeneity across clients in each round, and strikes a balance between inference accuracy, energy efficiency, and training time of FL. In addition, \system{} consists of a personalization module which brings consistency in the inference performance of multiple devices and users. Figure~\ref{fig:overview} presents an overview of \system{} and below we describe its main components.

\subsection{User-Centered FL Training} \label{sec:user_centered_fl}

In Figure~\ref{fig:swd}, we showed that the multi-device setup has a high degree of non-IID-ness due to the presence of both user and device heterogeneities. There are two extreme choices to reduce this non-IID-ness: 
\begin{enumerate}[label=(\alph*), leftmargin=*]
    \item \emph{Remove Device Heterogeneity}: we can select the same device from each user during a round of FL and ensure that there is no device heterogeneity across clients. For example, during Round 1, we can train the global model only on smartphone data from all users; during Round 2, we can train only on smartwatch data, and so on. This ensures that during each round of FL, only user-related heterogeneity is the source of non-IID-ness across clients. 
    \item \emph{Remove User Heterogeneity}: On the other extreme, one can consider removing user heterogeneity from each round of FL by training on each user sequentially, e.g., in Round 1, we train the global model on data from the devices of only user 1, and so on. 
\end{enumerate}

Both these extreme choices, however, are undesirable. In (a), the parameters of the global model can oscillate significantly between rounds due to the distribution shifts induced by the different devices used in each round and lead to poor convergence behavior. Similarly, (b) induces distribution shifts due to a drastic change in user characteristics between rounds, which could lead to poor convergence as shown in prior work~\cite{charles2021large}. {In addition, (b) significantly limits the number of clients participating in each round, slowing down the training process.}

\begin{figure}[t]
    \begin{minipage}{0.6\textwidth}
        \centering
        \includegraphics[width=0.5\linewidth]{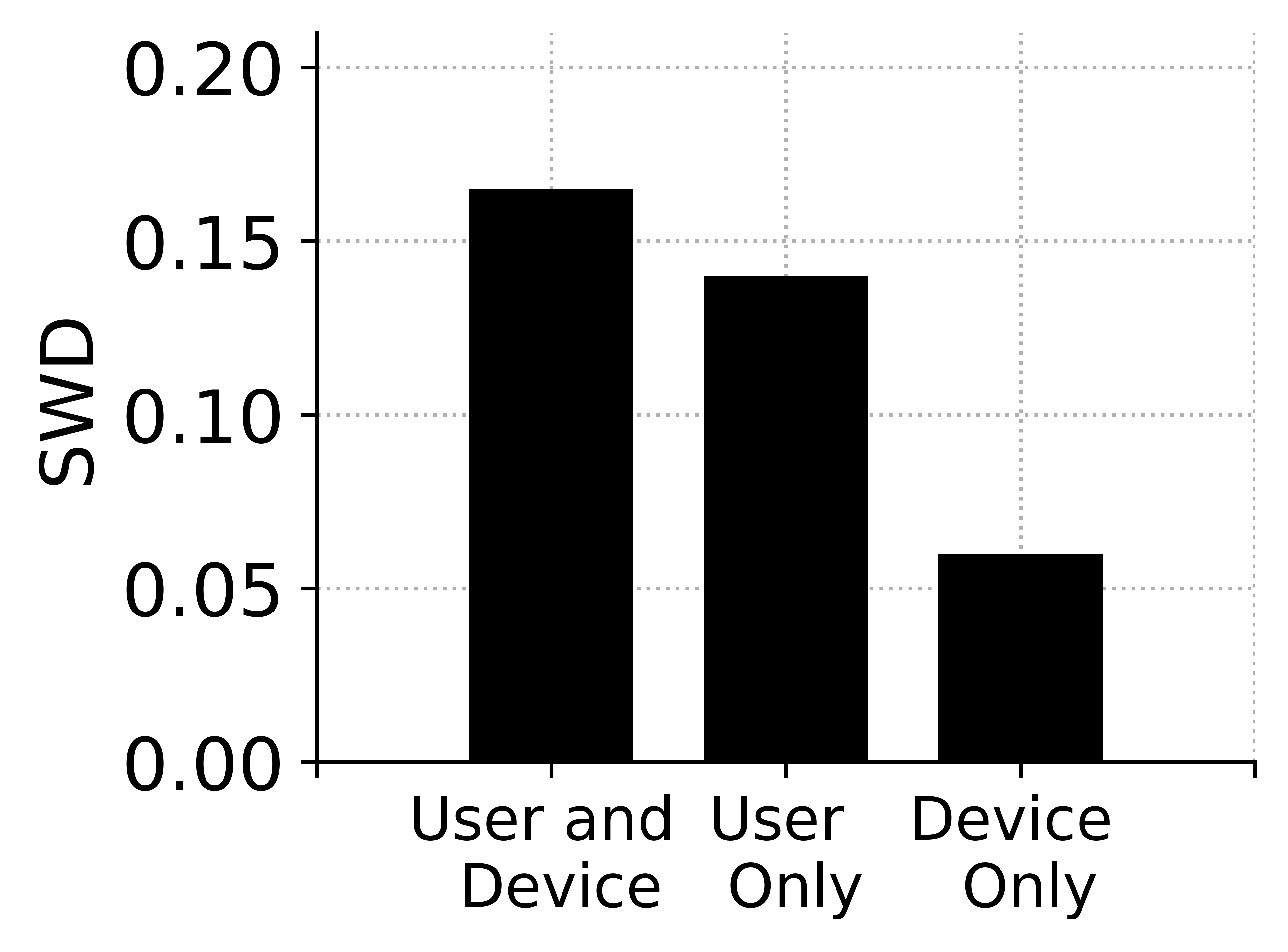}
    \captionof{figure}{Comparison of the impact of user heterogeneity vs. device heterogeneity on a multi-device FL setup. Same experiment setup as Figure~\ref{fig:swd} is used. To compute the mean user heterogeneity, we first fix a device (e.g., thigh-worn IMU) and calculate the mean pairwise SWD across all users for this device. This process is repeated for every device in the dataset to obtain the mean SWD due to user variations. To compute device heterogeneity, we fix a user (e.g., User 1) and calculate the mean pairwise SWD across all devices of this user. This process is repeated for every user in the dataset to obtain the mean SWD due to device variations.}
        \label{fig:swd_controlled}
    \end{minipage} \hfill
    \begin{minipage}{0.34\textwidth}
        \centering
        \includegraphics[width=1.0\linewidth]{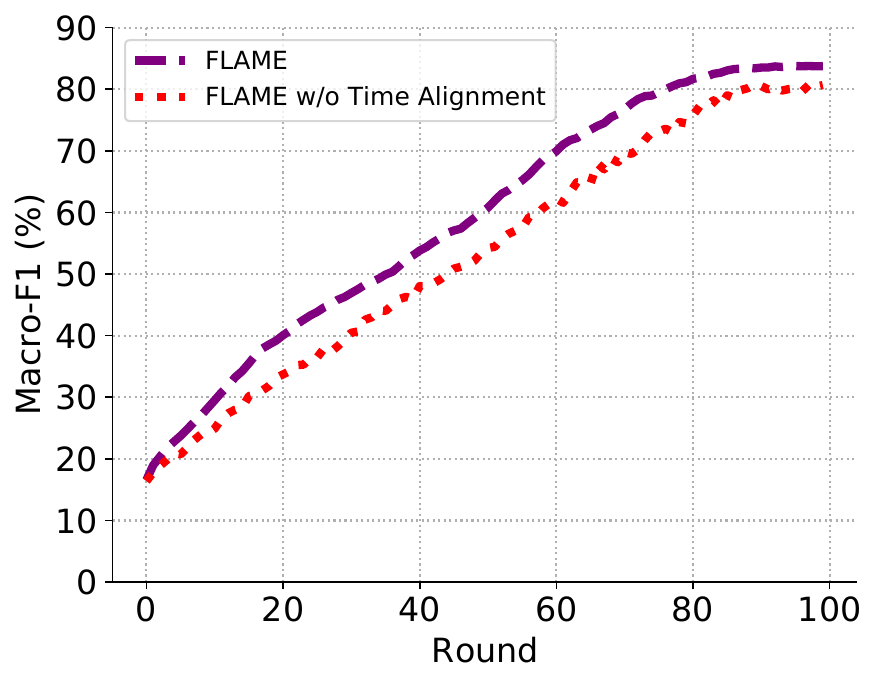}
    \vspace{-0.4cm}
    \captionof{figure}{Impact of enforcing time alignment in the MDE setup. \system{} w/o Time Alignment shows the worst case when such time alignment is not achievable, and \system{} uses randomly shuffled data instead of time alignment enforcement.}
    \label{fig:time_alignment_compare}
    \end{minipage}%
\end{figure}

\parjump{}
\noindent
Rather than completely eliminating either of these heterogeneity during training, we adopt a practical solution of minimizing their impact on the non-IID-ness of the setup. To motivate our solution, we first extend the experiment shown in Figure~\ref{fig:swd} and compare the impact of device vs. user heterogeneity on the non-IID-ness of a multi-device FL setup. Specifically, we compute the Sliced Wasserstein Distance (SWD) across clients in two scenarios: (a) When only \emph{User Heterogeneity} is present across clients, and (b) When only \emph{Device Heterogeneity} is present across clients. Our findings in Figure~\ref{fig:swd_controlled} show that the presence of user heterogeneity results in a higher SWD than device heterogeneity, and hence, they contribute more to the non-IID-ness of a FL setup. On the contrary, device heterogeneity alone causes less non-IID-ness because the devices in an MDE~(refer to \S\ref{sec:problem} for definition) collect sensor data in a synchronized and time-aligned manner. 

Based on this empirical insight, we propose a user-centered scheme for selecting FL clients in each round. Let $C$ be the total clients we wish to sample in each round, $N$ is the number of users in the \problem{} setup, $K$ is the number of devices owned by each user, and $\rho \in [1, K]$ is a parameter that balances user and device heterogeneity in the setup. In each round of FL, we select $\frac{C}{\rho} \subseteq N$ users. From each selected user, we choose $\rho$ number of time-aligned devices for federated training. If a user has less than $\rho$ available devices, we choose all available devices from the user and select more users until we sample $C$ total clients. As $\rho$ increases, more \emph{time-aligned} devices from each user are selected, while lesser number of unique users are sampled in a round. Both these factors contribute to reducing the overall non-IID-ness of the setup. We also note an interesting empirical finding that emerged from this design decision (see Figure~\ref{ablation-a}). It is observed that during FL, if we enforce the time-aligned devices of a user to follow the same order of iterating through their local datasets (as opposed to randomly shuffling their data), we can achieve faster convergence and higher overall accuracy. In practice, time alignment can be achieved through exchanging wall-clock time among co-located devices whose proximity could be known through various ways, e.g., bluetooth signal proximity. In more extreme scenarios where the data from one or more devices are missing or incomplete, previously collected data could be used instead; in the worst case, \system{} would perform as running with randomly shuffled data shown in Figure~\ref{fig:time_alignment_compare}.

\subsection{Accuracy- and Efficiency-Aware Device Selection} \label{sec:client_selection}
Having proposed a way to reduce the non-IID-ness of the \problem{} setup, we now need to strike a balance between the various performance metrics of a FL system, namely accuracy, convergence time, and energy efficiency. As explained above, we wish to select $C$ devices, spread over $\frac{C}{\rho}$ users (or more if there are users with less than $\rho$ available devices) in each round. Below we propose a sampling strategy based on the statistical, system, and time utility of each device.

\subsubsection{Statistical Utility}
Statistical utility refers to the utility of a device's data towards improving model performance. Following prior work by Lai et al.~\cite{oort}, we argue that a device which has a higher training loss on its local dataset should have a higher statistical utility. Our statistical utility metric for device $i$ in round $r$ is therefore defined as: 
\[stat(i, r) = |X_i| \sqrt{\frac{1}{|X_i|}\sum_{k\in{X_i}} \mathcal{L}_{r-1}(k)^2}\] where \(X_i\) is the set of locally stored training samples and \(\mathcal{L}_{r-1}(k)\) is the training loss of the global model from the previous round on sample $k$. Intuitively, this utility prioritizes the selection of devices which currently have a higher training loss over their local dataset, and hence could provide stronger gradients to update the global model. 
The statistical utility can range from 0 to $\infty$, depending on the loss function, e.g., categorical crossentropy loss. Higher value means higher utility.

\subsubsection{System Utility}
System utility is a device's utility toward improving system efficiency of FL. In this paper, we use energy drain as a representative metric of system utility, although it can be easily substituted with other metrics depending on the system requirements. Our system utility metric for device $i$ in round $r$ is defined as:
\[system(i, r) = \mathbbm{1} (drain^i_{r-1} < drain^i_{th}) \cdot log \left( \frac{drain^i_{th}}{drain^i_{r-1}}\right) \]
where $\mathbbm{1}$ denotes the indicator function. If device $i$'s accumulated energy drain due to federated training till the previous round ($drain^i_{r-1}$) exceeds a predefined energy drain threshold ($drain^i_{th}$), the device's system utility becomes zero. If not, devices with smaller accumulated energy drain have higher system utility. $drain^i_{th}$ is akin to an energy consumption budget for federated training and it could be predefined either by the system designer or the user. When a device runs out of its energy consumption budget, its system utility becomes 0 and it is no longer selected for training. The system utility ranges from 0 to $log(drain^i_{th})$, and higher value means higher utility.

\subsubsection{Time Utility}
Time utility encourages the selection of devices that complete their local training and communicate the model parameters to the server within a predefined time threshold. At the same time, it penalizes the slower devices and discourages their selection. We define time utility as: 
\[ time(i , r ) = 1 - \mathbbm{1}(t^i_{r-1} > T_{max}) \cdot \left( 1 - \alpha \cdot \frac{T_{max}}{t^i_{r-1}}  \right) \]
where $T_{max}$ is the desired duration of each round and $t^i_{r-1}$ is the time taken  by client $i$ to complete the previous round $(r-1)$. Specifically, \( t^i_{r-1} = t^i_{dl} + t^i_{ul} + t^i_{train} \) is the sum of the model weights download time ($t^i_{dl}$), upload time ($t^i_{ul}$), and the local training time ($t^i_{train}$) in round $(r-1)$. Our time utility metric penalizes clients whose total round completion time $t^i_{r-1}$ is higher than the threshold $T_{max}$ by a developer-defined factor $\alpha$, ranging from [0, 1].
Lower $\alpha$ will penalize slow clients more. For clients whose round completion time is less than $T_{max}$, time utility is always 1. 
Thus, the time utility ranges from 0 to 1, and higher value means higher utility.

\subsubsection{Overall Utility}
Finally, to strike a balance between accuracy, convergence time, and energy efficiency, we multiply the three utilities to obtain a unified utility metric for device $i$ in round $r$ is as follows: 
\vspace{-0.1em}
\[Util(i, r) = stat(i, r) \times system(i, r) \times time(i, r)\]

In \system{}, each device computes its unified utility and reports it to the server. In the next round, the server sorts all the devices by their unified utility value in a descending order, and select the top $C$ devices subject to the constraint that they are distributed across $\frac{C}{\rho}$ users (or more if there are users with less than $\rho$ available devices).

\subsection{Model Personalization}
\label{subsec.personalization}
Finally, to address the challenge of uneven accuracies across devices and users as shown in Figure \ref{fig:fedavg_user_variations}, \system{} personalizes~\cite{ditto} the global model to each device during training. Let $w_r$ denote the weights of the global model in a given round $r$, which are obtained by averaging the weights from each selected device. In addition to this global model, \system{} also trains a personal model on each device, which is trained using only the local data from the device. As this model is personalized to each device, it can be expected to result in higher inference accuracy as compared to the global model and avoid the challenge of uneven accuracies across devices.

Let $v_i$ denote the weights of the personal model for the device $i$, which are initialized randomly. Training the personal model follows an update rule below:
\[v_i = v_i - \eta(\nabla L_i(v_i) + \lambda(v_i-w^r)),\] where $\eta$ denotes the local learning rate, $\nabla L_i(v_i)$ is the gradient of the loss function on the local dataset, and $(v_i-w^r)$ is a regularization term that ensures that the weights of the personal model do not diverge too much from the global model. Finally, $\lambda$ is a hyperparameter that balances the local training objective and the regularization term; a higher $\lambda$ pushes the personal model towards the global model, while a smaller $\lambda$ encourages higher personalization. Please note that our model personalization approach is inspired by the work of Li et al.~\cite{ditto}; we do not claim novelty on the algorithm or its underlying theoretical foundations. Instead, we argue and empirically validate that the use of model personalization brings significant benefits for FL in a multi-device environment and addresses the challenge shown in Figure~\ref{fig:fedavg_user_variations}.

\parjump{}
\noindent
The algorithm for \system{} is summarized in Algorithm~\ref{alg:personalization} and illustrated in Figure~\ref{fig:overview}.
For every round $r$ until the final $R$-th round, the server selects a subset of devices, $\{C_r\}$, following the device selection strategy\footnote{Only in the first round, devices are sampled randomly as no prior information about their state is available.} described in \S\ref{sec:client_selection}. The server distributes the up-to-date global model weights, $w^r$ to each device $c \in \{C_r\}$. Every selected device updates the global model on its local data for $E$ epochs, generating new global model weights $w^r_c$. Along with the global model update, each device also updates its personal model following the personalization update rule above, generating $v^r_{c}$. The new global model weights $w^r_c$ of each client $c$ are sent back to the global server for aggregation using federated averaging algorithms. The personal model is not shared with the server.

\begin{algorithm}[t]
    \caption{\system{}}
    \begin{algorithmic}
        \For{$r = 0$ to $R-1$} 
            \State $C_r \gets $ A subset of clients are sampled   \Comment{Client selection (\S\ref{sec:client_selection}})
            \For{client $c \in C_r$ in parallel}
                    \For{local iteration $e = 0$ to $E$}
                    \State $w^{r}_{c} \gets $ ClientUpdate($c$, $w^r$)    \Comment{Global model update}
                    \State Send $w^r_c$ back to the global server for aggregation
                    \State $v^r_c \gets v^r_c - \eta(\nabla{\mathcal{L}_c(v^r_c)} + \lambda(v^r_c - w^r))$ \Comment{Device model update (\S\ref{subsec.personalization})}
                \EndFor
            \EndFor
            \State \(w^{r+1} \gets \sum^{\left|C_r\right|}_{i=1}\frac{n_i}{n}{w^r_i}\) \Comment{Server aggregates global model updates}
        \EndFor
    \end{algorithmic}
    \label{alg:personalization}
\end{algorithm}

\vspace{0.3cm}
\begin{tcolorbox}
\textbf{Takeaways:} \system{} employs a user-centered FL training approach in combination with a device selection scheme that balances accuracy, convergence time, and energy efficiency of FL. Further, the use of model personalization in \system{} is aimed at reducing the inconsistencies (or variance) in inference performance across devices.
\end{tcolorbox}

\section{Evaluation} \label{sec:evaluation}

We present a rigorous evaluation of \system{} on three multi-device HAR datasets by comparing the accuracy, time-efficiency and energy-efficiency of our approach against a number of FL baselines. Our key results are:

\begin{itemize}
    \item \system{} outperforms various federated learning baselines in terms of inference performance. \system{} achieves higher $F_1$ score (\revision{4.3-25.8}\%) than its baselines in all three datasets for both personalized and global models.
    
    \item \system{} results in less \emph{\dead{}} devices after 100 rounds of training for all three datasets. A device becomes `\dead{}' when its energy drain grows above a predefined drain threshold (see \S\ref{sec:eval_metrics}). Energy-aware device selection algorithm of \system{} leads to more balanced client distribution of the training efforts and reduces the final number of \dead{} devices by 1.02-$2.86\times$.
    
    \item \system{} speeds up convergence to a target accuracy. When we set the target accuracy to the lowest final accuracy among baselines, \system{} sped up the convergence up to \revision{$2.06\times$}.
    
    \item \system{} achieves the highest inference performance while maintaining system-side benefits over different device sampling strategy baselines and ablation settings.
\end{itemize}

\subsection{Datasets}

For our experiments, we use three multi-device datasets for human activity recognition: \textsc{Opportunity}, \textsc{RealWorld}, and \textsc{PAMAP2}. The characteristics of these datasets are described in Table~\ref{tab:dataset_information}.

\begin{table}[h]
	\centering
	\caption{Datasets used in the paper, along with their pre- and post-partitioning statistics.}
	\label{tab:dataset_information}
	\vspace{-1em}
	\begin{tabular}{lcccccc}
		\toprule
		\multirow{3}{4em}{Name} & \multirow{3}{4em}{\# Devices Per User}  &
		\multicolumn{2}{c}{Before Partitioning} & \multicolumn{2}{c}{After Partitioning} & \multirow{3}{4em}{Sampling Rate} \\
		\cmidrule{3-6}
		& & \# Users &    \begin{tabular}[c]{@{}c@{}}Average \# of training \\ samples per device\end{tabular} & \# Users & 
		\begin{tabular}[c]{@{}c@{}}Average \# of training \\ samples per device\\ \end{tabular} 
		  & \\
		\midrule
		RealWorld~\cite{sztyler2016onbody} & 7 & 15  & 1481 &  149 & 130  & 50Hz          \\
		Opportunity~\cite{Opportunity}     & 5           & 4          & 3847 & 28 & 356  & 30Hz          \\
		PAMAP2~\cite{pamap2}               & 3          & 8          & 1057 & 78  & 87 &100Hz         \\
		\bottomrule
	\end{tabular}
\end{table}

\parjump{}
\noindent
\textbf{RealWorld ~\cite{sztyler2016onbody}}: This dataset consists of data from 15 participants performing 8 locomotion activities: jumping, lying, standing, sitting, running, walking, climbing down, and climbing up. While performing the activities, 7 IMU-enabled devices were placed on the user's body at the following positions: head, chest, upper arm, waist, forearm, thigh, and shin. Accelerometer and gyroscope traces were recorded from the devices simultaneously at a sampling rate of 50 Hz.

\parjump{}
\noindent
\textbf{Opportunity ~\cite{Opportunity}}. This dataset consists of data collected from 4 participants performing activities of daily living with 17 on-body sensor devices. For our study, we used five devices deployed on back, left lower arm, right shoe, right upper arm, and left shoe, and we targeted to detect the mode of locomotion: \emph{stand, walk, sit, and lie}. Accelerometer and gyroscope traces were recorded from the devices simultaneously at a sampling rate of 30 Hz.

\parjump{}
\noindent
\textbf{PAMAP2 ~\cite{pamap2}}. This dataset includes data recorded from 9 subjects performing 18 different activities. As 6 were optional activities, we only used 12 activities among them: ascending stairs, cycling, descending stairs, ironing, lying, nordic walking, rope jumping, running, sitting, standing, vacuum cleaning, and walking. Users were instrumented with IMUs placed at 3 different body positions: head, chest, ankle; and accelerometer and gyroscope data were sampled from the devices simultaneously at a sampling rate of 100 Hz. We chose this dataset because in addition to locomotion activities, PAMAP2 also contains examples of activities of daily living (ADL) such as ironing, vacuum cleaning, and rope jumping. 

\parjump{}
\noindent
Based on prior works, we segment the accelerometer and gyroscope data in time windows of 3 seconds for RealWorld and 2 seconds for Opportunity and PAMAP2 datasets without any overlap~\cite{chang2020systematic, liono2016optimal}. 

\subsection{Experiment Testbed}
\label{subsec:testbed}
To perform a realistic evaluation of a FL system, we need an experiment testbed that simulates the characteristics of real-world federated learning. One of the key motivations behind FL is that local clients have an \emph{insufficient amount of training data} to learn a good prediction model, and hence they collaborate with other clients to learn a shared prediction model. Unfortunately, some of the prior work on FL with HAR data disregards this assumption; for example, Yu et al.~\cite{yu2021fedhar} assume between 3000 to 5000 seconds of labeled data on each client for the RealWorld dataset. This raises two concerns: firstly, it is infeasible for users to label several hours of accelerometer and gyroscope data on their devices. Secondly, and perhaps more critically, if each client has a large number of labeled data samples, they may not even need collaborative training algorithms such as FL to train prediction models.

Hence, for a robust evaluation of FL systems, we need to find the ways to partitioning existing HAR datasets to make them appropriate for federated settings. Next, we present a novel data partitioning scheme that can take any multi-device HAR dataset and distribute it over a large number of client devices. \emph{We are in the process of open-sourcing the source code of our partitioning algorithm as well the federated HAR datasets for reproducibility}.

\subsubsection{Class-Based Data Partitioning for New User Generation}
A realistic FL testbed should have a large number of users each owning a small number of labeled data samples. To this end, our data partitioning scheme generates new users by partitioning the data of existing users in the dataset. This scheme preserves the multi-device nature of the \problem{} setting and performs \textit{class-based} data partitioning. 

\begin{figure}
    \centering
    \includegraphics[width=0.9\linewidth]{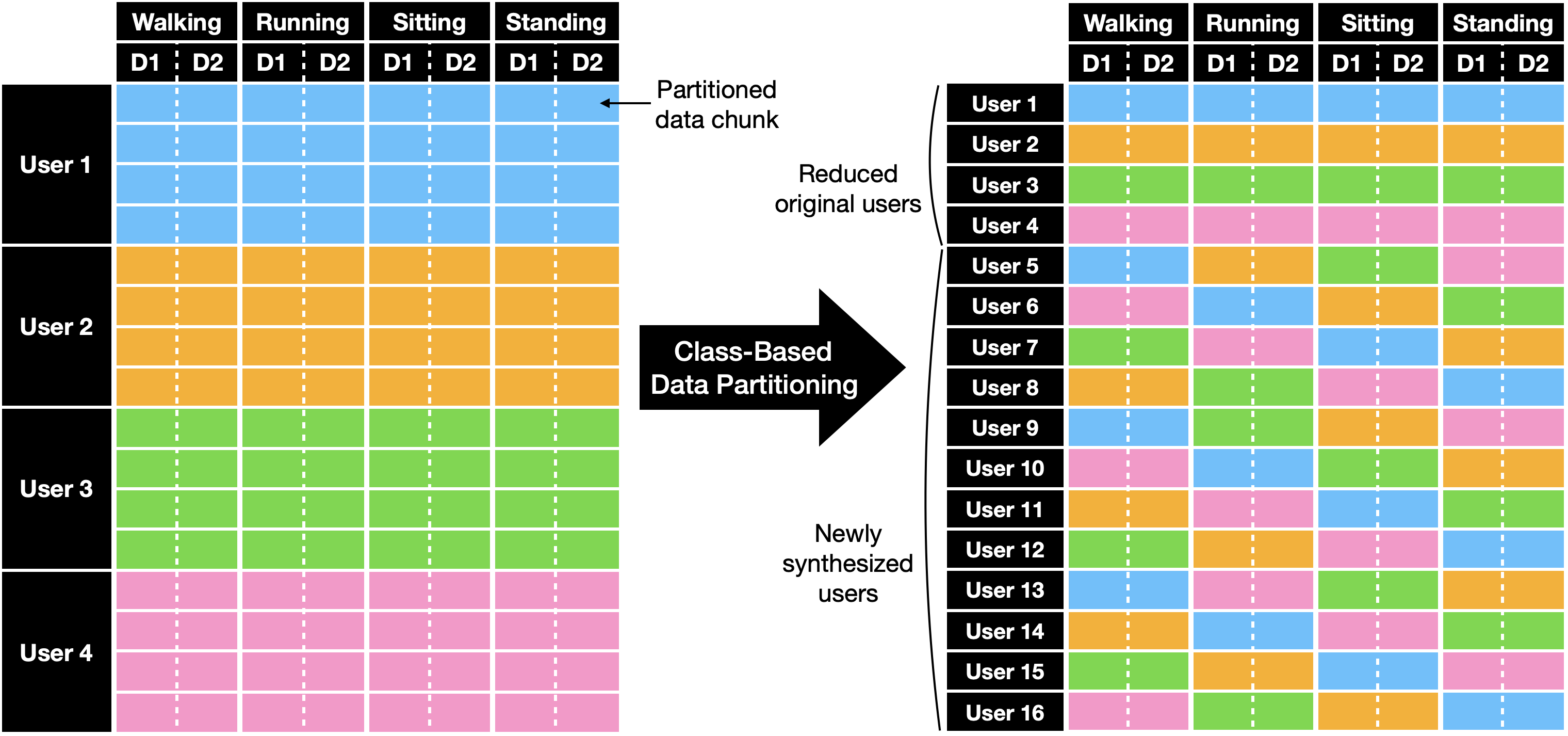}
    \caption{Class-based data partitioning for a realistic federated learning multi-device user generation. An example of partitioning 4 users, 2 devices per user~(D1 and D2), and 4 activity classes~(Walking, Running, Sitting, and Standing) generates 12 new users with distinct profiles.}
    \label{fig:data_partitioning}
\end{figure}

Figure~\ref{fig:data_partitioning} demonstrates an example of partitioning a dataset with $N=4$ users, $L=4$ classes~(walking, running, sitting, standing), and $D=2$ devices per user~(D1, D2). Assume that we wish to partition this dataset across $N'=16$ users in total (i.e., create 12 new users). We first divide the data samples of each of the existing users into $N'/N$ chunks. The original $N=4$ users keep one chunk of their data and rest of the chunks are distributed to create 12 ($N' - N$) new users. Each new user is created by mixing different classes from the original users in a distinct combination; for example, the user 5 in Figure~\ref{fig:data_partitioning} contains `Walking' class from User 1, `Running' from User 2, `Sitting' from User 3, and `Standing' from User 4. 

The main advantages of our partitioning scheme are that (a) it preserves the characteristics of the multi-device  environment (MDE), in that the synthesized users borrow different activity classes from different original users; however within each activity class, all devices from the original users remain intact. This ensures that our partitioned dataset does not violate the definition of an MDE provided in \S\ref{sec:problem}. (b) We ensure that a new user gets data for an activity (e.g., running) from \emph{only} one of the original users. This maintains the integrity of activity classes and avoids scenarios where a synthesized user may have samples for the same activity from different original users. 

Please refer to Table~\ref{tab:dataset_information} for details on the partitioned datasets. As an example, the RealWorld dataset has 1481 training samples (74 minutes of data) per device from 15 users before partitioning. After partitioning, it changes to 149 users with 130 training samples (6 mins of data) per device.

\begin{table}[]
\small
\caption{Embedded processor profiles with training time and energy consumption per round for training an HAR model on the partitioned RealWorld dataset. Similar profiles are created for the other two datasets.}
\vspace{-0.3cm}
\label{tab:energy_profiles}
\begin{tabular}{@{}ccccc@{}}
\toprule
\textbf{Name}                                                      & \textbf{Processor Type} & \textbf{Release Year} & \textbf{\begin{tabular}[c]{@{}c@{}}Training \\ time per round \\ (in seconds)\end{tabular}} & \textbf{\begin{tabular}[c]{@{}c@{}}Energy consumption\\ per round \\ (in Joules)\end{tabular}} \\ \midrule
\begin{tabular}[c]{@{}c@{}}Raspberry Pi 4\\ (Model B)\end{tabular} & CPU                     & 2019                  & 38.18                                                                                       & 69.87                                                                                          \\ \midrule
\multirow{2}{*}{Jetson Nano}                                       & CPU                     & 2019                  & 50.31                                                                                       & 27.3                                                                                           \\ 
                                                                   & GPU                     & 2019                  & 33.10                                                                                       & 22.5                                                                                           \\ \midrule
\multirow{2}{*}{Jetson Xavier NX}                                  & CPU                     & 2019                  & 23.12                                                                                       & 15.5                                                                                           \\
                                                                   & GPU                     & 2019                  & 16.11                                                                                       & 13.7                                                                                           \\ \midrule
\multirow{2}{*}{Jetson AGX Xavier}                                 & CPU                     & 2018                  & 16.0                                                                                        & 8.85                                                                                           \\ 
                                                                   & GPU                     & 2018                  & 11.11                                                                                       & 7.36                                                                                           \\ \midrule
\multirow{2}{*}{Jetson TX2}                                        & CPU                     & 2016                  & 42.79                                                                                       & 128.9                                                                                          \\
                                                                   & GPU                     & 2016                  & 28.73                                                                                       & 87.3                                                                                           \\ \bottomrule
                                                                   \vspace{-0.8cm}
\end{tabular}
\end{table}

\subsubsection{Energy Drain Profiles}
Our FL testbed also incorporates the energy drain profiles of modern embedded processors, thus allowing us to quantify the energy drain associated with training models on real client devices. To this end, we measure the energy consumption of training an HAR model on 9 different processors as shown in Table~\ref{tab:energy_profiles}. These devices are chosen because of their compatibility with Python that enable us to federate TensorFlow model training on them. For energy profiling, we choose the same training configuration of the HAR model (i.e., network architecture, learning rate, local epochs, dataset) that is used in our end-to-end FL experiments (see \S\ref{sec:hparams}). We run the training for five rounds and measure the mean training time per round and the mean energy consumption on each processor separately. In total, this results in 9 realistic device profiles for each dataset. Finally, in our large-scale FL evaluation, each client in the FL setup (obtained after data partitioning) is randomly assigned one of the 9 device profiles. When a client participates in a training round, it experiences energy drain according to its assigned profile. \revision{An additional round of profiling for training without personalization revealed that the energy budget is 70\% of the numbers reported in Table~\ref{tab:energy_profiles}. This value was used for baselines that do not employ per-round personalization.}

We note that a recent work called FedScale~\cite{lai2021fedscale} took a similar approach of creating an experiment testbed for FL; however they did not include any simulation of realistic energy consumption on edge devices.

\subsubsection{Realistic Network Bandwidths}
Finally, to simulate varying network bandwidth across users, our testbed assigns a realistic network bandwidth to each user participating in FL. We collect the data on average download/upload speeds on mobile devices in different countries from SpeedTest~\cite{opensignal}, and each user in the FL system is randomly assigned a download/upload speed from this dataset. All devices of a user share the same download/upload speed.

\subsection{Baselines} \label{sec:baselines}

We use two federated learning baselines to compare against \system{}: FedAvg~\cite{BrendanMcMahan2017} and Ditto~\cite{ditto}. Unlike \system{}, FedAvg considers each device in our problem setup as a separate client and ignores the association between different devices of a user. We use FedAvg as a baseline as it is the original algorithm proposed for federated learning. In each round of training, FedAvg randomly samples $C$ devices and averages their models on the server. Ditto~\cite{ditto} is a state-of-the-art personalization extension of FedAvg that also includes the training of personalized models. Further, we compare \system{} against Oort~\cite{oort} which is an extension of FedAvg with heterogeneity-aware device selection policies. We use Oort as a baseline for being the state-of-the-art device selection policy that considers both statistical and time utility. We also evaluate \textsc{Power-of-choice} ($\pi_{pow-d}$)~\cite{cho2020client} and its computation-efficient variant ($\pi_{cpow-d}$) to compare \system{} to another device selection strategy focusing on statistical utility ($\pi_{pow-d}$) and both statistical and computational utility ($\pi_{cpow-d}$). As our work operates under the paradigm of supervised FL, we do not compare it against any unsupervised or semi-supervised FL baselines~\cite{yu2021fedhar, Bettini2021}.

\subsection{Implementation Details and Hyperparameters} \label{sec:hparams}
For our experiments, we use the DeepConvLSTM model architecture proposed for HAR~\cite{deepconvlstm}. Our FL training setup is implemented in Tensorflow using the Flower framework~\cite{beutel2020flower}. We use the TF HParams API for hyperparameter tuning and arrived at the following training hyperparameters: \{\system{} learning rate = $1e^{-3}$, batch size = 32, optimizer = Adam, rounds = 100, local epochs = 20 for RealWorld and PAMAP2 and 10 for Opportunity\}. 
For other variables in \system{}, we use personalization factor $\lambda$ = 1.0, device sampling ratio = 0.5, $\alpha$ for time utility = 0.5, $\alpha$ for Oort's time utility = 2.0, and $b$ for $\pi_{cpow-d}$'s minibatch size = 50\% number of samples.

\subsection{Evaluation Metrics} \label{sec:eval_metrics}
We use a 80-20 train-test split of each dataset; the training set was used for updating the personal and global models and the testing set was used for evaluation. Please note that unlike the more robust $k$-fold cross validation or leave-one-user-out evaluation, we opted for a simpler train-test split evaluation. This is done due to the high costs (both monetary and environmental) associated with federated learning experiments. For instance, each experiment on the RealWorld dataset requires federated training on 1043 devices (149 users x 7 devices per user) for 20 local epochs and 100 rounds, which in total consumes around 104 GPU-hours on Nvidia V100 GPUs. Hence, we decided not to repeat this experiment for multiple folds to reduce the monetary, and more importantly the environmental costs of training.

We compare \system{} against the baselines on three metrics: inference performance, energy drain, and convergence speed. For the inference performance, we report the macro $F_1$ score which is considered a good performance metric for imbalanced datasets~\cite{plotz2021applying}. For energy drain, we report the number of devices that become `\dead{}' for training in each round. A device becomes `\dead{}' when its energy drain grows above a predefined drain threshold. Assuming a 3000mAH battery, we used a conservative drain threshold of 10\% of the battery capacity.

\subsection{Results} \label{sec:results}

In this section, we present our results comparing \system{} against the baselines in a number of experiment settings.

\begin{figure}[t]
\subfloat[RealWorld dataset]{\label{f1-a}\includegraphics[width=0.33\linewidth]{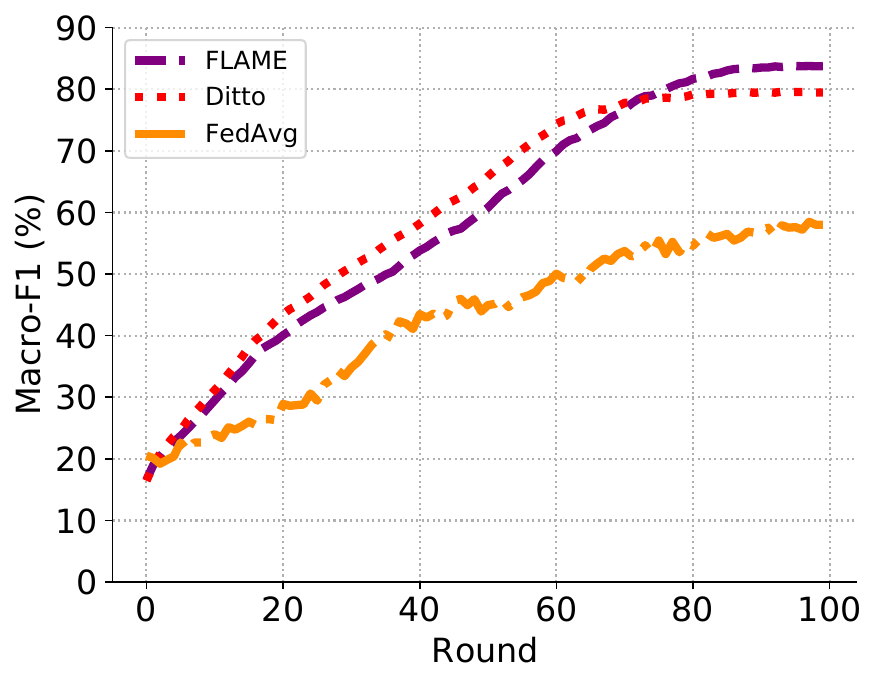}}\hfill
\subfloat[Opportunity dataset]{\label{f1-b}\includegraphics[width=0.33\linewidth]{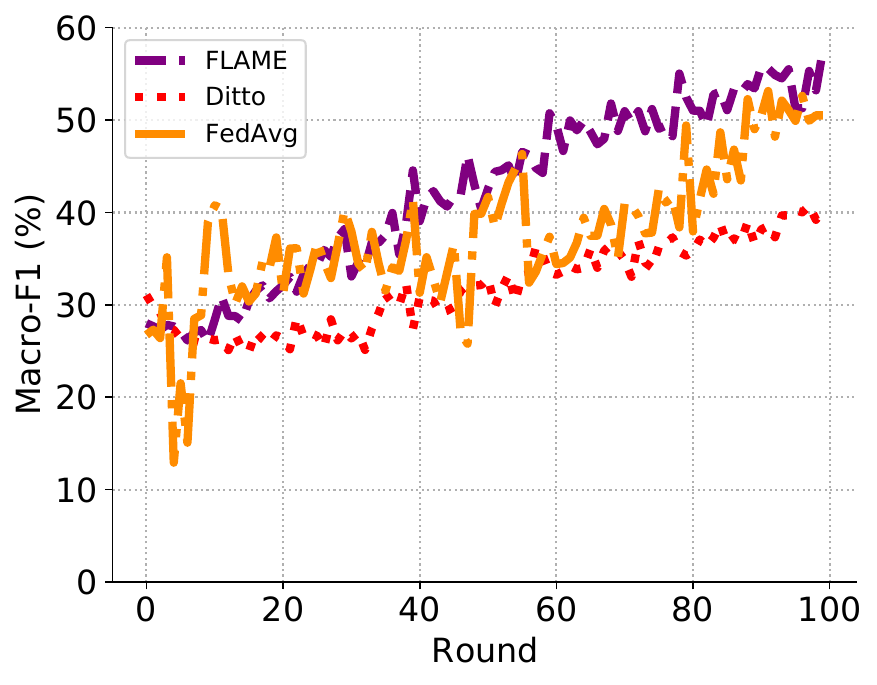}}\hfill
\subfloat[PAMAP2 dataset]{\label{f1-c}\includegraphics[width=0.33\linewidth]{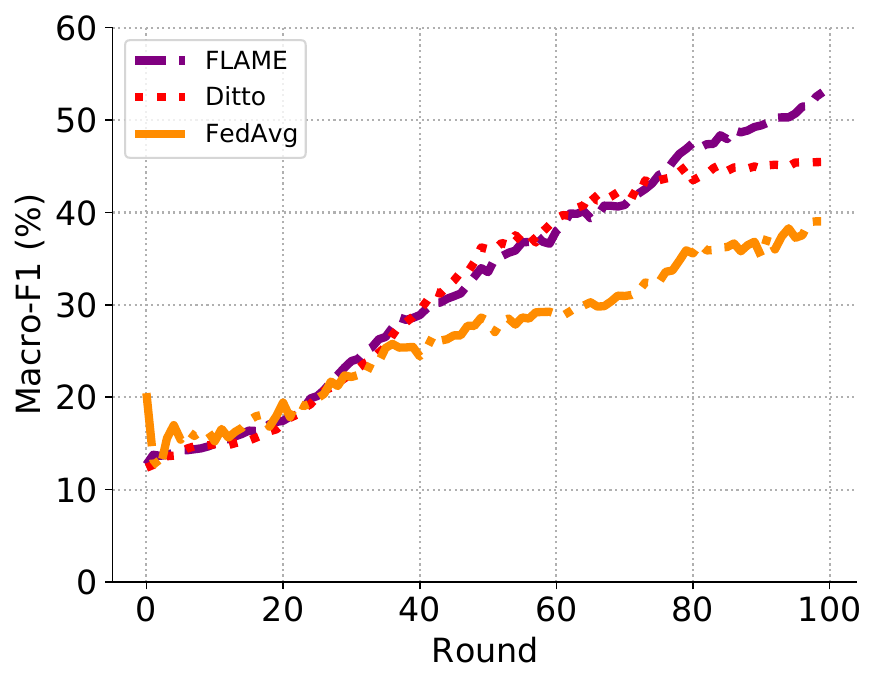}}\hfill
\caption{Round-to-F1 score plot.}
\label{fig:round_to_accuracy}

\subfloat[RealWorld dataset]{\label{dead-a}\includegraphics[width=0.33\linewidth]{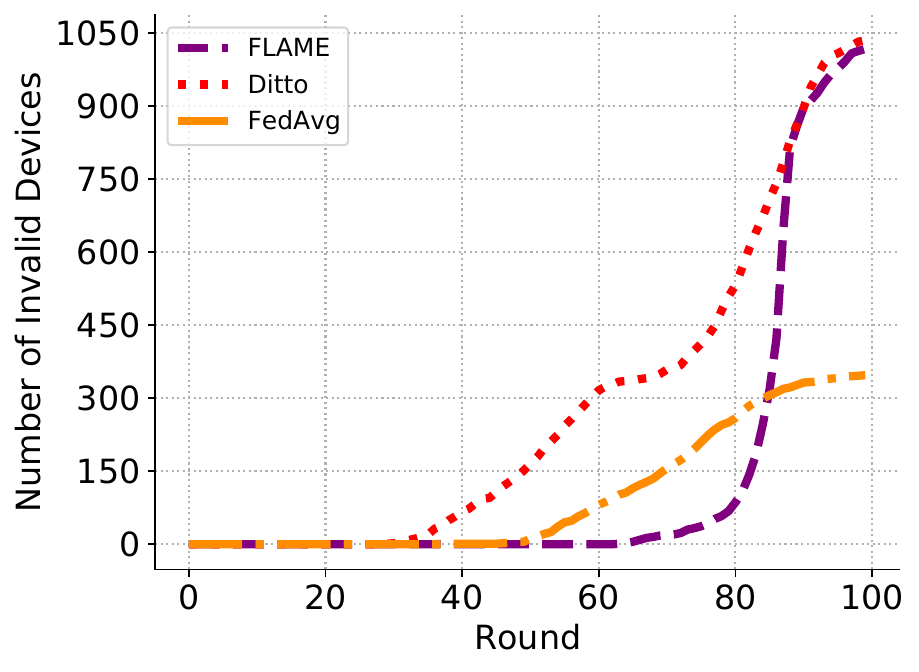}}\hfill 
\subfloat[Opportunity dataset]{\label{dead-b}\includegraphics[width=0.33\linewidth]{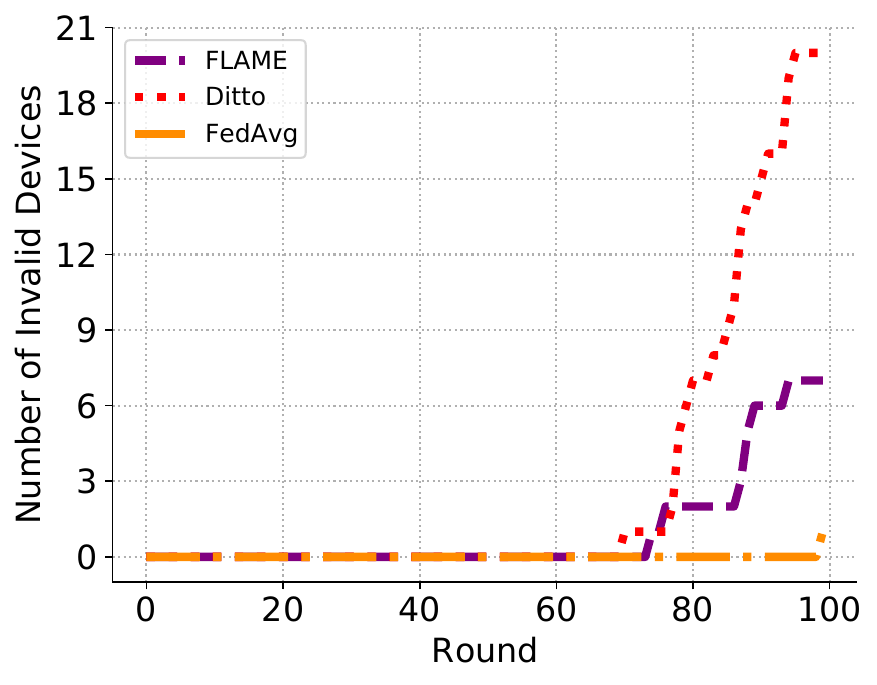}}\hfill 
\subfloat[PAMAP2 dataset]{\label{dead-c}\includegraphics[width=0.33\linewidth]{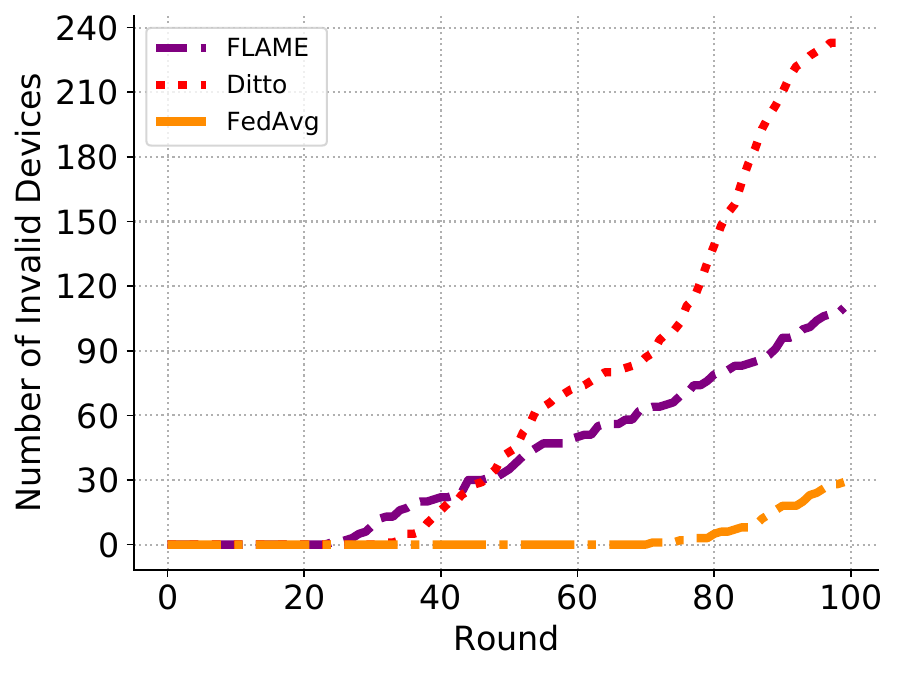}}\hfill 
\caption{Number of \dead{} devices over rounds.}
\label{fig:dead_devices}
\end{figure}

\begin{table}[t]
\centering
\caption{End-to-end performance of \system{} compared to Ditto and FedAvg baselines. The final model $F_1$ score of FedAvg's global model is used as the target $F_1$ score for speedup calculation. The best performance in each column is marked in bold. Device model results for FedAvg is marked `N/A' as FedAvg does not train a device model. \revision{Local Finetuning applies local finetuning to the final global model of FedAvg after 100 rounds of training.}
Ditto does not reach the target $F_1$ score with its device model in Opportunity dataset, hence its speedup is marked `-'.}
\vspace{-0.2cm}
\begin{tabular}{@{}llccccc@{}}
\toprule
\multirow{2}{4em}{Dataset} & \multirow{2}{4em}{Algorithm} & \multicolumn{2}{c}{Macro-$F_1$ Score} &  \multirow{2}{4em}{Target $F_1$ score} & \multicolumn{2}{c}{Speedup}\\
		\cmidrule{3-4}\cmidrule{6-7} & & Device & Global &   & Device & Global \\ 
		\midrule
\multirow{4}{*}{RealWorld} 
 & \system{} & \textbf{83.8\%} & 52.6\% & \multirow{4}{*}{58.0\%} & 2.06$\times$ & - \\ 
 & Ditto & 79.5\% & 51.7\%& &\textbf{2.42$\times$ }& - \\ 
 & FedAvg & N/A & \textbf{58.0\%} & & N/A & 1.00$\times$ \\
 & Local Finetuning & \revision{75.6\%} & \textbf{58.0\%} & & \revision{1.00$\times$} & 1.00$\times$ \\
\midrule
\multirow{4}{*}{Opportunity} 
 & \system{} &  \textbf{57.5\%} & 48.8\%  & \multirow{4}{*}{50.5\%} & \textbf{1.80$\times$} &  \textbf{1.49$\times$}\\
 & Ditto & 40.3\% & 48.8\% & & - & - \\
 & FedAvg & N/A & \textbf{50.5\%} & & N/A & 1.00$\times$ \\
 & Local Finetuning & \revision{53.2\%} & \textbf{50.5\%} & & \revision{1.00$\times$} & 1.00$\times$ \\
\midrule
\multirow{3}{*}{PAMAP2}
 & \system{} & \textbf{53.0\%} & \textbf{39.7\%}  & \multirow{3}{*}{39.0\%} & 1.58$\times$ &  1.05$\times$ \\ 
 & Ditto & 45.5\% & 22.7\% & & \textbf{1.63$\times$} & \textbf{1.08$\times$} \\
 & FedAvg & N/A & 39.0\% & & N/A & 1.00$\times$\\
 & Local Finetuning & \revision{51.6\%} & 39.0\% & & \revision{1.00$\times$} & 1.00$\times$ \\
\midrule
\end{tabular}
\label{tab:perf_summary}
\end{table}

\parjump{}
\noindent
\textbf{\system{} improves the test accuracy of FL models.} 
Figure~\ref{fig:round_to_accuracy} shows the plot for macro $F_1$ score obtained per round, and Table~\ref{tab:perf_summary} summarizes the final $F_1$ score for personalized device models and global models averaged over all the devices in the dataset. In all three datasets, \system{} achieved the highest $F_1$ score over Ditto and FedAvg after 100 rounds of federated training. 
\revision{Since FedAvg does not train a device model, we added a Local Finetuning baseline in Table~\ref{tab:perf_summary} which applies local finetuning~(cf. \S3.3.3 of Kairouz et al. \cite{flsurvey}) to the final FedAvg global model after 100 training rounds. Refer to \S\ref{sup:local_finetuning} for more details about local finetuning experiments.}
Interestingly, we observe that in RealWorld and PAMAP2 datasets, \system{} and Ditto show similar inference performance at start; however, once the number of \dead{} devices start increasing in Ditto (see Figure \ref{fig:dead_devices}), its performance saturates due to a smaller pool of available devices to train on. This shows how energy-aware device selection could benefit not only the energy consumption but also the prediction performance of FL algorithms.

\parjump{}
\noindent
\textbf{\system{} distributes the FL workload across devices in an energy-aware manner, resulting in less \dead{} devices after training than Ditto.} Figure~\ref{fig:dead_devices} shows the number of \dead{} devices over rounds. It is important to note that due to local model training, devices will inevitably run out of the energy budgets assigned to them, more so if the budgets are conservative. Hence, an ideal FL algorithm should: a) result in lesser number of \dead{} devices after training, and b) distribute the workload equitably across devices such that the onset of \dead{} devices is delayed during training. From Figure~\ref{fig:dead_devices}, we can observe that \system{} has less \dead{} devices than Ditto for all three datasets, and it succeeds in delaying the onset of \dead{} devices for RealWorld and Opportunity datasets. FedAvg achieves the least number of \dead{} devices but with incomparable macro $F_1$ scores.

\begin{figure}[t]
    \begin{minipage}{\textwidth}
        \centering
        \vspace{-0.4cm}
        \includegraphics[width=0.35\linewidth]{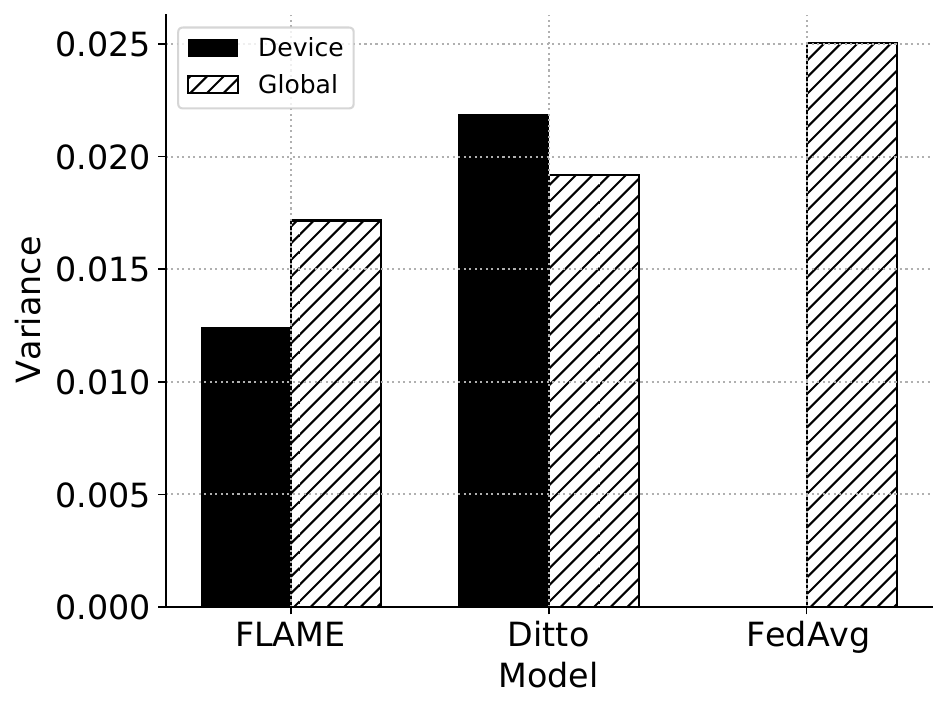}
        \vspace{-0.4cm}
        \captionof{figure}{Variance of device and global model inference $F_1$ score across each user's devices, averaged over users.}
        \vspace{-0.4cm}
        \label{fig:variance}
    \end{minipage}
\end{figure}

\parjump{}
\noindent
\textbf{\system{} lowers the variance in $F_1$ score across devices of the same user.} As shown in Figure \ref{fig:fedavg_user_variations}, FL models often do not generalize well to different devices, and may result in uneven accuracies across devices of the same user. Figure~\ref{fig:variance} shows the variance in the macro $F_1$ score of personalized and global models across devices of the same user. We first compute the variance in macro $F_1$ across all devices of each user. The average of these variances over all users is reported in Figure~\ref{fig:variance}. We observe that both device and global models trained with \system{} lower the across-device variance, compared to Ditto and FedAvg, thereby leading to more equitable performance across devices.

\parjump{}
\noindent
\textbf{Convergence time of \system{}.} Table~\ref{tab:perf_summary} shows the convergence time speedup for \system{}. As the two baselines do not achieve the same accuracy as \system{}, we use the final accuracy of the least accurate baseline (FedAvg) to measure the convergence speedup. In other words, we measure the time taken by \system{} and Ditto to achieve the same accuracy as FedAvg. Our results show that \system{} achieves speedups for the device models as they converge faster due to personalization with 2.06$\times$, 1.80$\times$, and 1.58$\times$ speedups, respectively for RealWorld, Opportunity, and PAMAP2 datasets.
\system{}'s global model shows slight speedups with Opportunity and PAMAP2 datasets with 1.49$\times$ and 1.05$\times$ speedups respectively, but it does not reach the FedAvg accuracy with the RealWorld dataset in 100 rounds. We attribute this to the slower energy depletion in FedAvg as it involves no personalization.

\begin{figure}[t]
\centering
\subfloat[Device model round-to-accuracy]{\label{sampling-a}\includegraphics[width=0.4\linewidth]{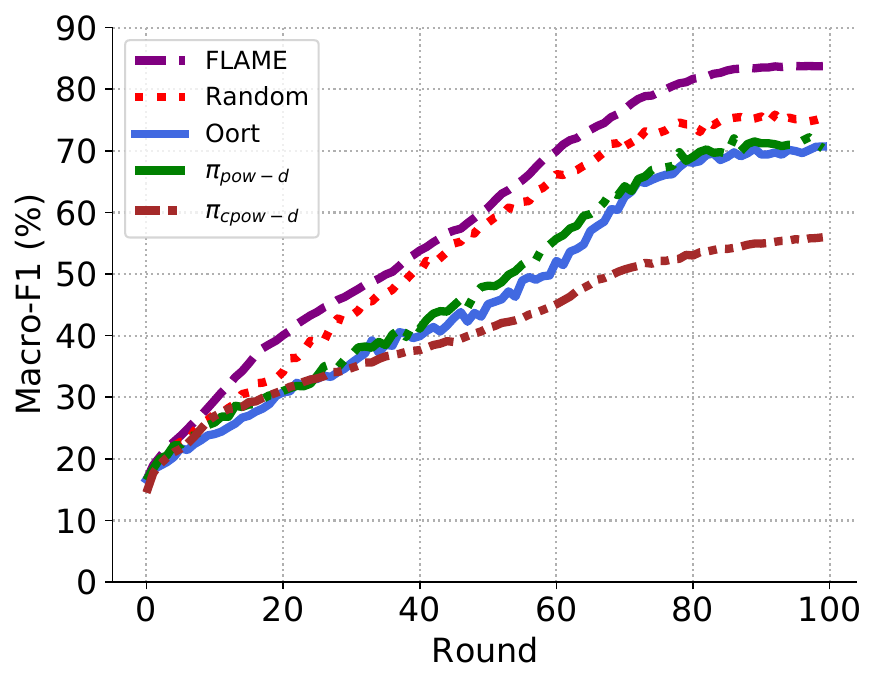}}
\subfloat[\Dead{} devices]{\label{sampling-b}\includegraphics[width=0.4\linewidth]{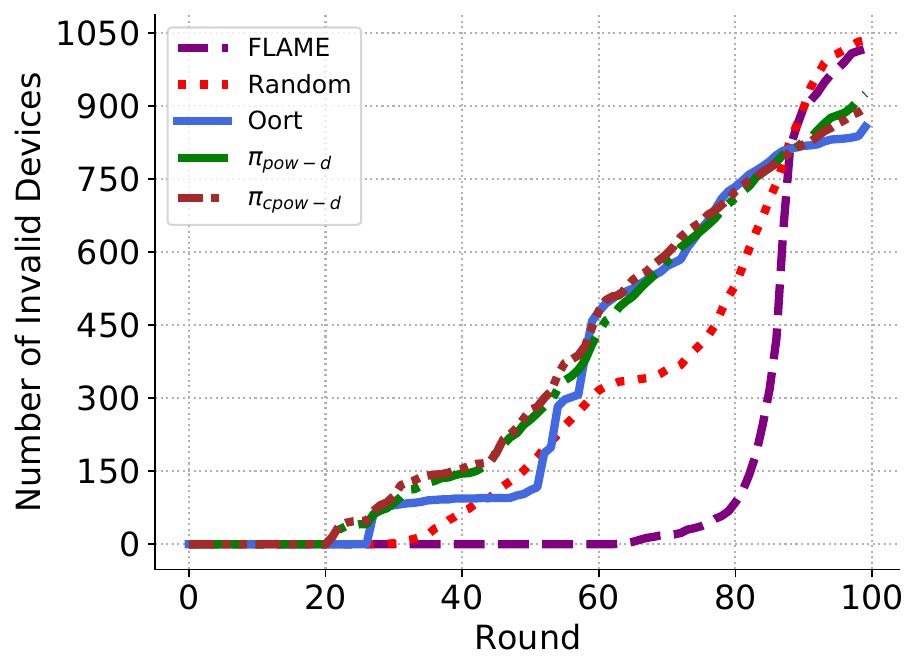}} 
\caption{Comparison between different device sampling strategies using the RealWorld dataset.}
\label{fig:device_selection}
\end{figure}

\subsection{Analysis of \system{}} \label{sec:analysis_flame}
In this section, we present ablation studies and a detailed analysis on the constituent algorithms of \system{}. 

\parjump{}
\noindent
\textbf{\system{} outperforms baseline device selection strategies.} We compare \system{}'s unified device selection strategy~(in \S\ref{sec:client_selection}) against the following baselines:
\begin{itemize}
    \item \textbf{Random}: devices are randomly selected for training.
    \item \textbf{Oort}: device selection based on Oort~\cite{oort} which combines statistical utility and a different time utility defined as $(\frac{T}{t_i})^{\mathbbm{1}(T<t_i)\times\alpha}$, favoring clients with shorter training time $t_i$ when $t_i$ is greater than the preferred threshold $T$. 
    \item \textbf{\textsc{Power-of-choice}} ($\pi_{pow-d}$): device selection based on \textsc{Power-of-choice}~\cite{cho2020client} which considers statistical utility by selecting clients with higher training loss.
    \item \textbf{\textsc{Power-of-choice} computation-efficient variant} ($\pi_{cpow-d}$): a variant of \textsc{Power-of-choice} which uses only a minibatch of $b$ samples to calculate the training loss.
\end{itemize}
Figure~\ref{fig:device_selection} illustrates how the personalized model $F_1$ score and number of \dead{} devices changes over rounds for all sampling strategies. \system{} achieves the highest inference $F_1$ score using the personalized model when compared to the other device sampling strategies. Oort has an overall less number of \dead{} devices; however, the onset of \dead{} devices starts early in round 27, which hampers its convergence. \system{} on the contrary manages to distribute the training workload across devices, and the onset of \dead{} devices only starts around round 70.

\subsubsection{Ablation Study}
\begin{figure}[t]
\centering
\subfloat[Round-to-accuracy]{\label{ablation-a}\includegraphics[width=0.4\linewidth]{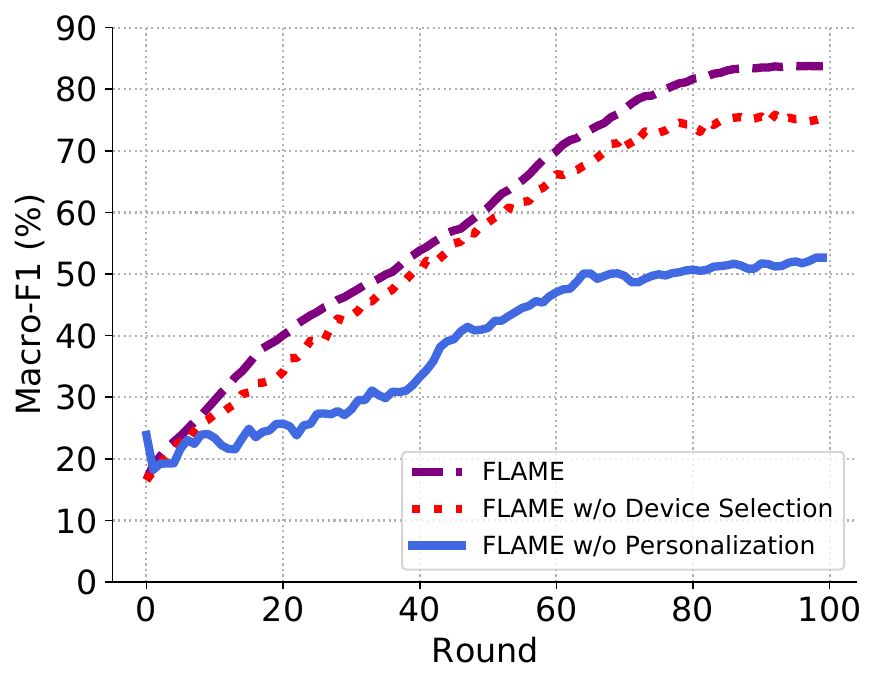}}
\subfloat[\Dead{} devices]{\label{ablation-b}\includegraphics[width=0.4\linewidth]{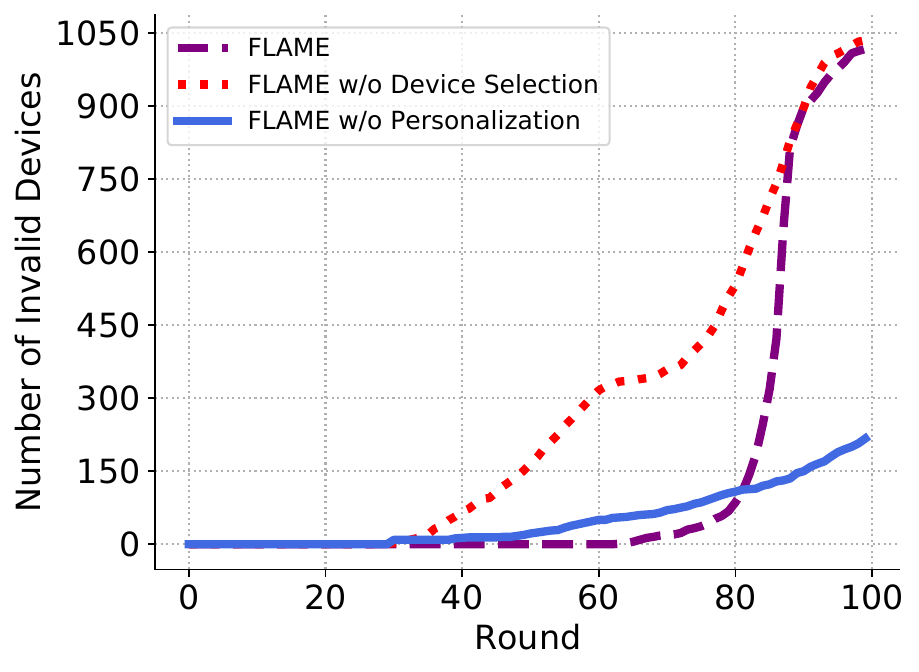}}
\caption{Ablation study using the RealWorld dataset.}
\label{fig:rw_ablation}
\end{figure}

Figure~\ref{fig:rw_ablation} shows our ablation study results, comparing \system{} with ablations of \system{} without our device selection strategy~(i.e., random sampling)
and without personalization. We observe that \system{} achieves the highest $F_1$ score over the ablation baselines while achieving comparable system performance.

\subsubsection{Varying the Number of Clients Selected in Each Round}
\begin{figure}[t]
\centering
\subfloat[Device model round-to-accuracy]{\label{selected-a}\includegraphics[width=0.4\linewidth]{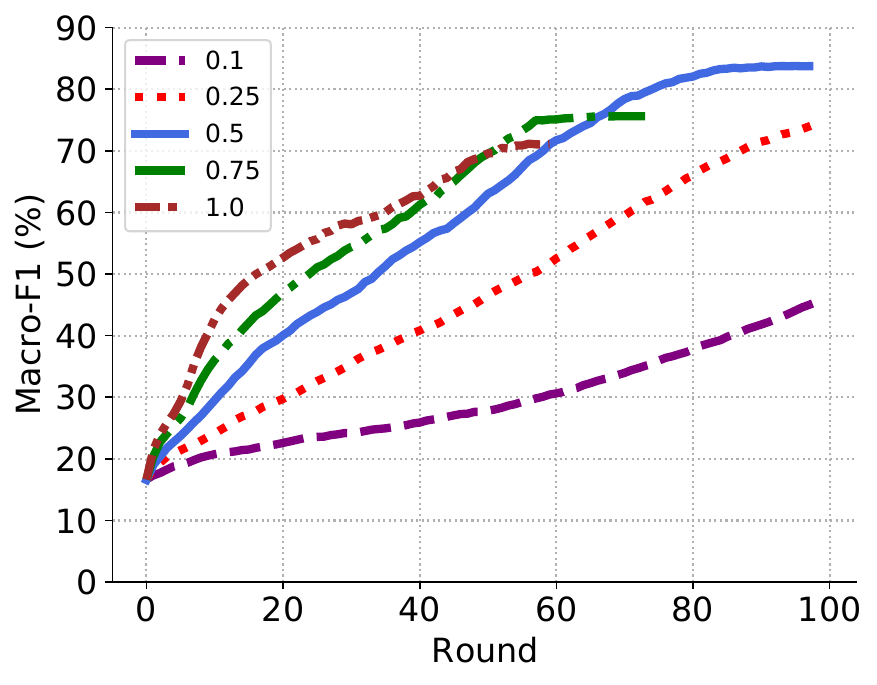}}
\subfloat[Number of \dead{} devices]{\label{selected-b}\includegraphics[width=0.4\linewidth]{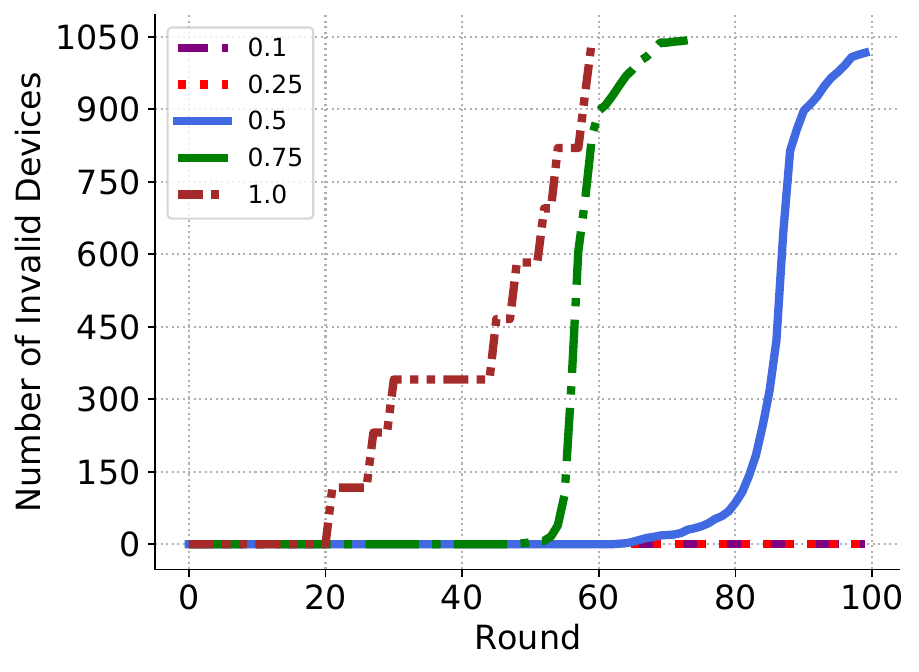}}
\caption{Varying the number of devices selected in each round using the RealWorld dataset.}
\label{fig:device_count}
\end{figure}

In our main experiments, we used a client sampling ratio of 0.5, that is 50\% of the total devices were sampled in each round of FL. We now present an experiment on the RealWorld dataset measuring the impact of varying the number of clients sampled in each round, from 0.1, 0.25, 0.5, 0.75, 1.0 fraction of total clients. Figure~\ref{fig:device_count} shows the per-round $F_1$ score for personalized models and the number of \dead{} devices during training. We observe that sampling more clients (fractions: 0.5, 0.75, or 1.0) leads to faster convergence to higher $F_1$ score than sample fractions of 0.25 or 0.1.  At the same time, we see diminishing returns with oversampling; for example, sample fraction of 0.5 performs better than 0.75 and 1.0. Such diminishing returns associated with oversampling have been shown in prior works as well~\cite{charles2021large}. Another explanation for this result is that sampling higher fraction of clients in each round depletes the energy of participating devices faster, leading to more \dead{} devices and reducing the overall $F_1$ score.

\begin{figure}[t]
\subfloat[Device 1]{\label{util-a}\includegraphics[width=0.4\linewidth]{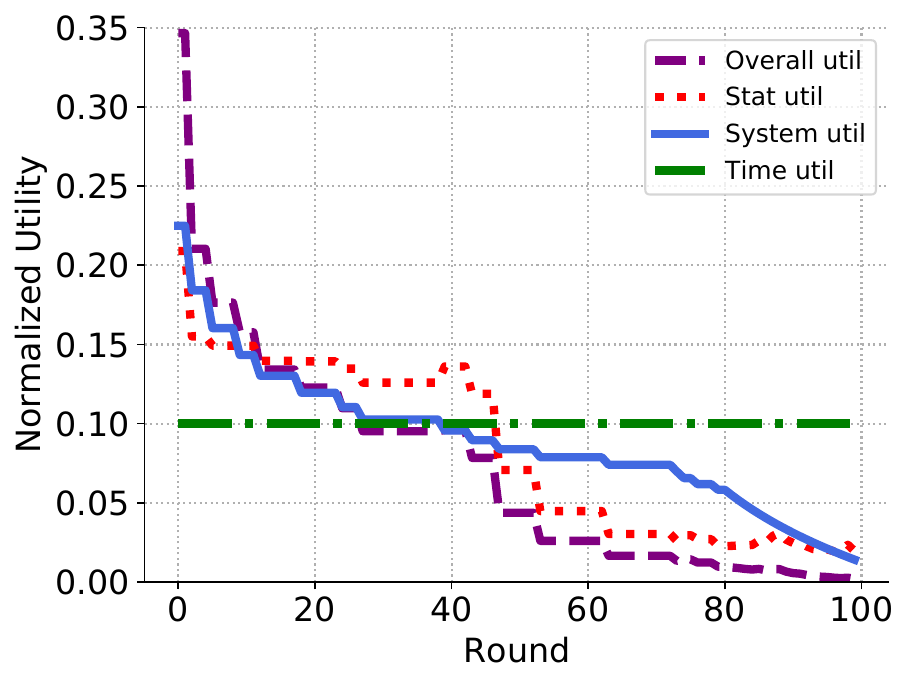}}
\subfloat[Device 2]{\label{util-b}\includegraphics[width=0.4\linewidth]{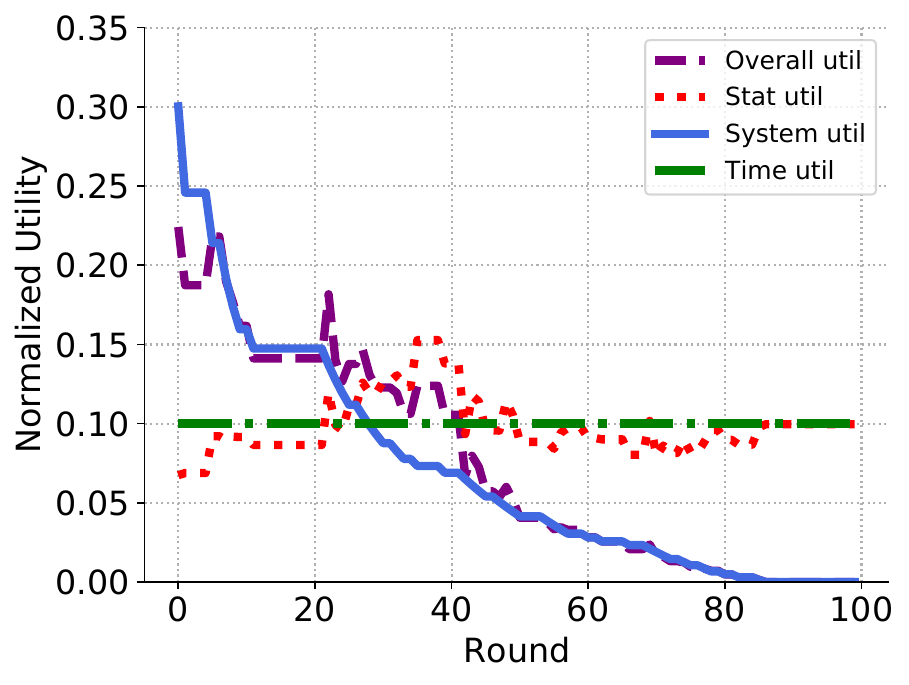}}\par 
\caption{Illustration of utility variation over rounds for two devices in the RealWorld dataset.}
\label{fig:utility_variation}
\end{figure}

\subsubsection{Illustration of the Device Selection Strategy}
Figure~\ref{fig:utility_variation} illustrates how the three utility components~(Statistical, System, Time) and the overall utility change over 100 rounds for two devices: \emph{Device 1}, where our selection strategy manages to balance statistical and systems utility over time and \emph{Device 2}, where it falters to some extent. The plot is normalized to depict all utility components in the same plot. 

Device 1 starts with a high statistical and systems utility and is frequently sampled by our strategy until round 23. This can be verified by the constant reduction in the system utility for the device. After round 23, the statistical utility of Device 1 becomes low (i.e., it does not provide enough useful gradients to the latest global model) and its system utility also drops. As such, the device is sampled less frequently for training. This ensures that the energy budget of the device is not exhausted early, and the device manages to contribute to FL until the last round. On the contrary, Device 2 is sampled frequently throughout the training process due to its high overall utility caused by a high systems utility (until round 15) and high statistical utility (round 20 onward). As a result, this device exhausts its energy budget around round 85.

\subsubsection{Generalizability of the Learned Classifiers}
Finally, we evaluate the generalizability of the learned global and device models on new users and new devices of the same user through leave-one-user-out~(LOUO) and leave-one-device-out~(LODO) experiments, on the RealWorld dataset. For the LOUO experiments, we randomly select three users~(S1, S2, and S11) to be left out. In each LOUO experiment, one of the three users is excluded from dataset generation. Therefore, no newly generated users inherit data from the left out user in training, and the global and device models trained with all other users are tested on this held-out user. In the LODO experiments, one of the seven devices in the RealWorld dataset is held-out and the models are trained on the remaining devices. 

\begin{figure}[t]
    \subfloat[Bootstrapping new users]{\label{generalizability-a}\includegraphics[width=0.4\linewidth]{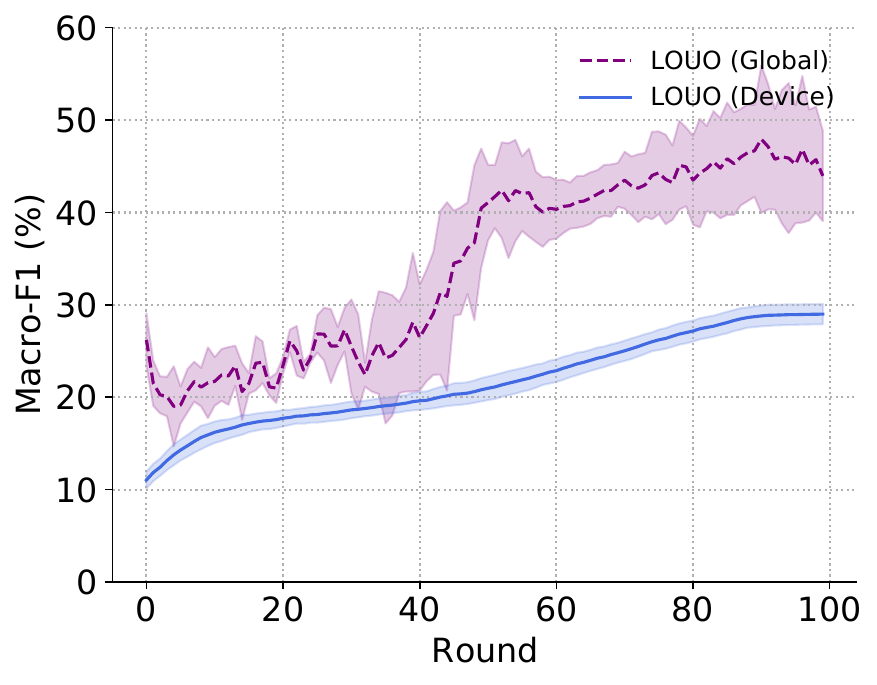}}
    \subfloat[Bootstrapping new devices]{\label{generalizability-b}\includegraphics[width=0.4\linewidth]{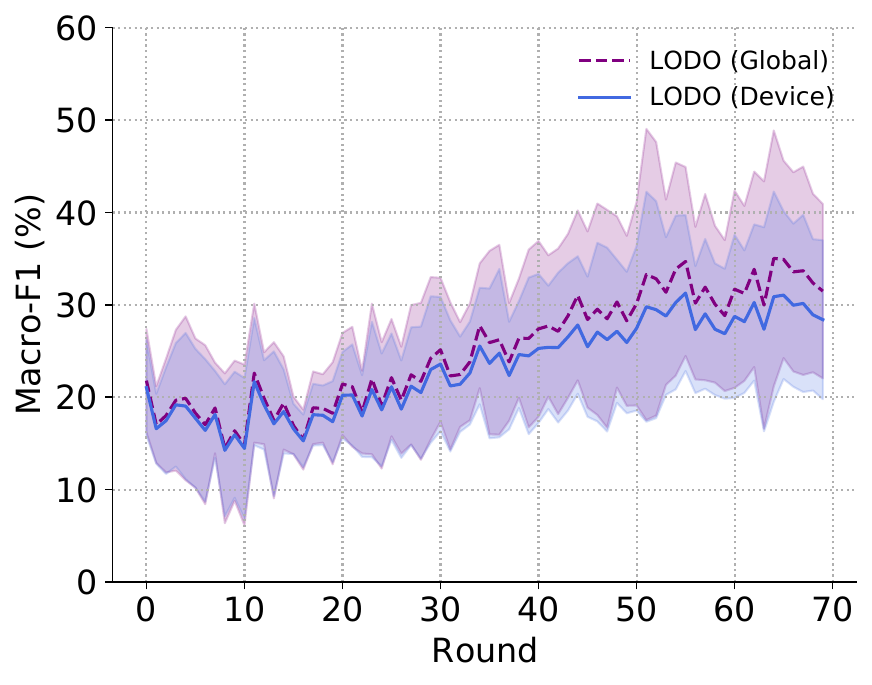}} 
    \caption{Macro-$F_1$ scores of learned device and global models tested on unseen users and unseen devices. This depicts better generalizability of global models than device models.}
    \label{fig:generalizability}
\end{figure}

Figure~\ref{generalizability-a} and \ref{generalizability-b} report the mean and standard deviation of $F_1$ scores over 100 rounds for the three LOUO and seven LODO experiments. In both cases, we observe that the global models achieve higher test accuracies for unseen users and unseen devices, which suggests that the global FL models generalize better than the personal models. This result highlights the need for maintaining both personal and global models during FL; while personal models offer higher inference $F_1$ score on each device, global models could be used to bootstrap HAR inference for new users and new devices that may appear in the future. Eventually, we expect that the new users or devices will also participate in FL training and learn a personalized model using \system{}.

\subsubsection{Changing the End of the Experiment}
\begin{figure}[t]
\centering
\subfloat[Round-to-accuracy]{\label{accuracy_end}\includegraphics[width=0.4\linewidth]{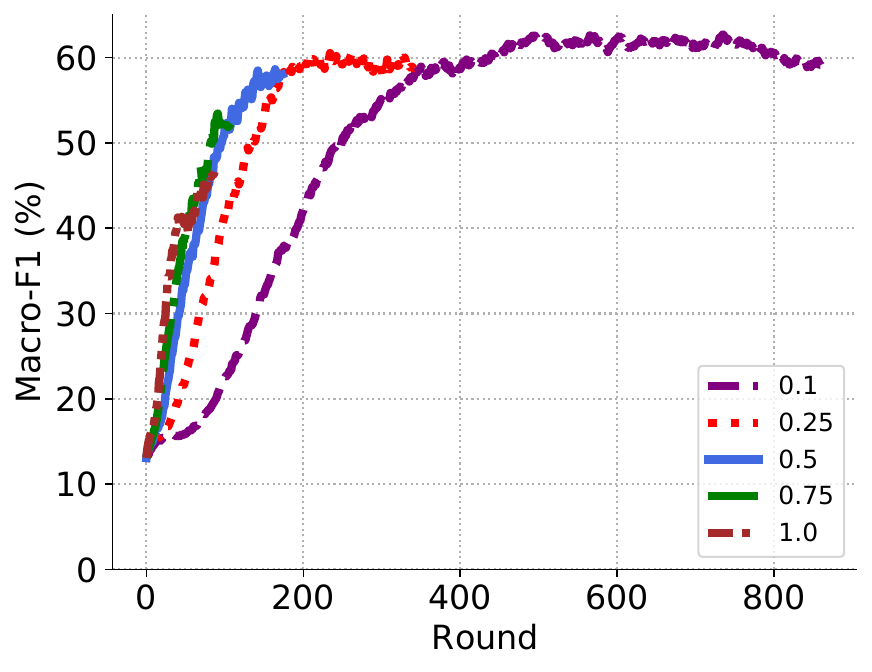}}
\subfloat[\Dead{} devices]{\label{dead_device_end}\includegraphics[width=0.4\linewidth]{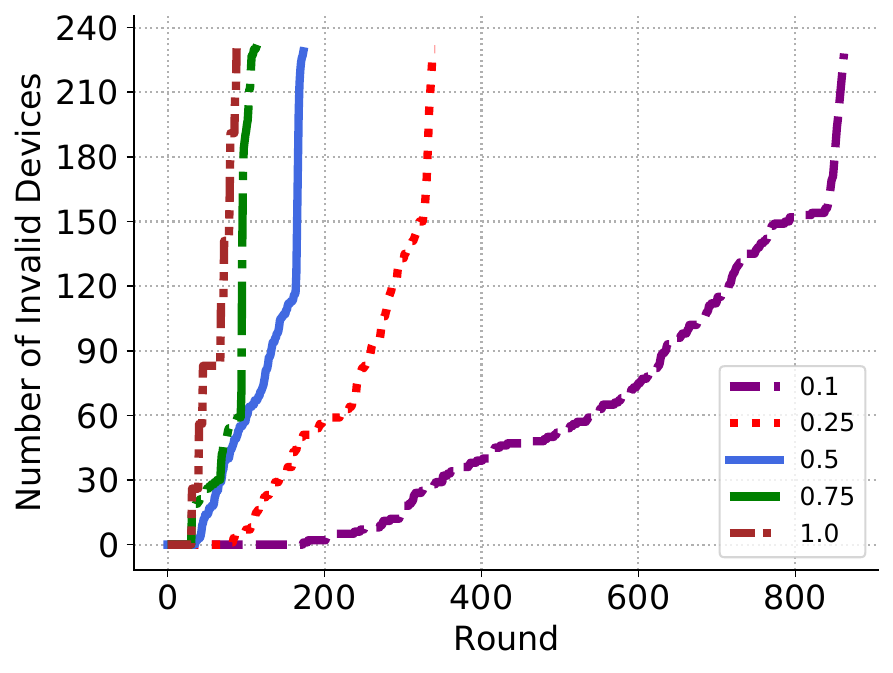}}
\caption{\revision{Varying the number of devices selected in each round using the PAMAP2 dataset. The experiment ran until 230 devices went invalid.}}
\label{fig:round_end}
\end{figure}

\revision{Figure~\ref{fig:round_end} shows the result of running \system{} with different sampling ratios using the PAMAP2 dataset in Figure~\ref{fig:device_count} when the end of the experiment is modified to where 230 devices become invalid instead of fixed to 100 rounds. The round in which 230 devices become invalid varies from round 89 to round 864.
It shows that runs with the sampling ratio of 0.25 and 0.1 have slower convergence rates. Up to round 175 in which 230 devices run out with the sampling ratio 0.5, both models with 0.25 and 0.1 achieve significantly less macro-F1 scores. However in later rounds, they reach comparable or surpass the final accuracy of 0.5 as they continue more rounds of training while exhausting less numbers of devices. This shows that a more sustainable policy with a lower sampling ratio could achieve higher macro-F1 score in the end, given sufficient time for training.}

\section{Discussion and Limitations}
\label{sec:limitations}
In this section, we elaborate on the limitations of \system{} and discuss future work on this topic.  

\subsubsection*{Generalizability of \system{}}
While the evaluation of \system{} was done on HAR tasks based on inertial sensing data, the core ideas of \system{} can be applied to train machine learning models in other types of multi-device systems, such as multiple microphones collecting a user's speech in a smart home or multiple cameras assessing driving behavior at a traffic junction. Further, our FL testbed is highly adaptable and can be easily re-purposed to partition and scale any small-scale sensor datasets for FL evaluations.

\subsubsection*{The Use of Energy and Latency Profiles}
We created energy and inference latency profiles of various embedded devices, and assigned those profiles to the FL clients in our evaluation. This methodology of simulating the energy drain and latency behavior of clients can assist FL researchers in quantifying the resource consumption of their algorithms in a quasi-realistic setup. That said, currently we have only profiled nine embedded processors specifically for HAR models. In order to scale this approach, more fine-grained profiles should be developed that can quantify the resource consumption on client devices based on the number of mathematical operations needed during local model training. 

\subsubsection*{Model Optimization and Averaging Algorithms}
In this paper, we did not explore the use of model optimization, which is an effective solution to speed up and reduce the resource consumption of local training~\cite{caldas2019expanding, han2020adaptive, jiang2019model}. Moreover, we only focused on FedAvg as the averaging algorithm for local parameters; the choice of averaging algorithm is an active area of research in FL and newer approaches are being proposed to deal with non-IID-ness during averaging~\cite{li2021fedbn, fedprox}. Although an in-depth study of averaging algorithms or model optimization techniques is out of the scope of this paper, we believe they can be easily incorporated with \system{} in future works. 

\subsubsection*{The Reliance on Labeled Data}
In its current form, \system{} operates under the paradigm of supervised federated learning~\cite{mcmahan2017communication}. We assumed that each device will have a small yet sufficient number of labeled data instances to drive federated training as well as model personalization. This assumption comes along with the well-known drawbacks of supervised learning on sensor data, e.g., the challenges associated with data labeling. We note that unsupervised~\cite{pmlr-v162-lubana22a} and semi-supervised FL~\cite{Bettini2021, sarkar2021grafehty, yu2021fedhar} are active research areas in ML and future works can investigate them in the context of multi-device systems. 

\subsubsection*{Similarity-Based Client Clustering}
An alternative approach to reducing non-IID-ness in sampled clients is to perform similarity-based clustering on their data and sample clients with higher similarity. We did not adopt this approach because in order to get a robust measure of distribution similarity, each client needs to share either raw data or features extracted from the data with the server. This violates the privacy of the personal data on each device. However, recent works have shown that one can get an approximation of client similarity by computing statistical measures such as the KL-Divergence on model parameters~\cite{clusterfl, dennis2021heterogeneity}. These approaches are more private, and could be extended to multi-device settings in future work.

\subsubsection*{Data Privacy}
\system{} does not share any raw data or features across devices, or between the devices and the server during FL. Hence, it provides some level of privacy protection. However, recent works in FL have shown that model parameters exchanged during training could leak the raw data~\cite{boenisch2021curious}. This remains an open challenge with \system{} as well. Furthermore, \system{} assumes tha the server has shared knowledge about user-device pairs in MDEs. Techniques such as differential privacy~\cite{wei2020federated} are being employed in FL literature to address these challenges, and they could be applied to \system{} in future work.        

\subsubsection*{Extension to Multi-Modal Data}
We evaluated \system{} with multi-device datasets containing only inertial sensor data from accelerometers and gyroscopes. Future works should investigate the performance of \system{} with other modalities such as audio, vision, physiological data, as well as in a multi-modal case where different local devices possess different modalities of data.

\subsubsection*{Extension of the Class-Based Partitioning Scheme}
We see two potential limitations and extensions of our proposed partitioning scheme. Firstly, it is possible that the FL model, instead of learning the true activity classification problem, learns to classify samples based on the original users they came from. Our empirical results in \S\ref{app:classification} highlight that this does not happen in our HAR tasks, however one must remain careful and analyze such inadvertent information leakage in the latent space. Should such leakage happen, one can consider doing data augmentation to the activity samples before they are assigned to the synthesized users. Secondly, our partitioning scheme ensures that the synthesized users have all activity classes present in them. This can be extended to an even more challenging scenario, where newly created users have only a subset of activities present. To this end, partitioning approaches based on Dirichlet splits~\cite{dirchsplit} could be employed.

\section{Related Work} \label{sec:related_work}

\subsubsection*{Overview of Research Challenges in Federated Learning}
Despite being a relatively new area in machine learning, FL has seen tremendous interest from both machine learning and systems community~\cite{lisurvey}. FL can be categorized into two types depending on the scale and nature of federation: \emph{cross-silo} and \emph{cross-device}~\cite{flsurvey}. In \emph{cross-silo} settings, large organizations collaborate to train a model, often using data center infrastructure and large local datasets. The more constrained scenario is that of \emph{cross-device} FL wherein clients are generally thousands of mobile or wearable devices with limited computational capabilities and relatively small amount of local data. Our work focuses on \emph{cross-device} FL. Recent works on this topic have investigated the issues of statistical and systems heterogeneity~\cite{fedprox, scaffold, dirchsplit, noniidfl, flgn}; achieving scalability, privacy, and fairness with respect to participating clients~\cite{charles2021large, flsecureagg, fairallocation, ditto, flmultitask}; and improving communication efficiency~\cite{feddyn, flcomms, fedopt, oort}. Our work builds upon this line of research, albeit in the context of multi-device FL systems. We tackle the statistical and system heterogeneity in multi-device FL using a user-centered training approach which promotes the participation of time-aligned devices in each round of FL. 

\subsubsection*{Client Selection in Federated Learning}
Client selection strategies in FL have been proposed to optimize statistical and system utilities based on clients' data and resource information~\cite{oort, Cho2020, Nishio2019, cho2020client, dennis2021heterogeneity}.
\revision{Cho et al.~\cite{cho2020client} present} \textsc{Power-of-choice}\revision{, which} samples highest loss clients for client selection in each round\revision{, and its} communication- and computation-efficient variants.
Oort~\cite{oort} also considers both statistical utility and time utility by selecting higher loss clients who can process samples within the preferred round duration.
$k$-FED~\cite{dennis2021heterogeneity} clusters data into $k$ heterogeneous clusters, which can be used as a prior to heterogeneity-aware client selection.
FedCS~\cite{Nishio2019} focuses on maximizing the number of clients to participate in a round based on the upload and update time, in scenarios where clients upload their model updates sequentially one by one. UCB-CS~\cite{Cho2020} introduces applying the discounted UCB algorithm, a multi-armed bandit algorithm, to balance the exploration-exploitation tradeoff communication-efficient client selection in FL.
We build upon the literature on client selection in FL and propose a unified client selection metric to balance model accuracy, training speed, and energy efficiency in multi-device FL.

\subsubsection*{Personalization in Federated Learning}
To address statistical heterogeneity and fairness challenges across FL clients, researchers have studied the personalization of local models to client characteristics. Mocha~\cite{flmultitask}, Ditto~\cite{ditto}, IFCA~\cite{ghosh2020efficient}, and ClusterFL~\cite{clusterfl} leverage multi-task learning to model structural relationships in distributed data and use the similarities for personalization. GraFeHTy~\cite{sarkar2021grafehty} applies a Graph Convolution Network with graph representations of activity inter-relatedness. APFL~\cite{deng2020adaptive} and MAPPER~\cite{mansour2020three} perform personalization through mixing the local and global models via model interpolation. FedDL~\cite{tu2021feddl} and FjORD~\cite{fjord} dynamically adapt model layers and model size for heterogeneous devices to participate in FL. Some studies further employ self-supervised or semi-supervised learning methods to utilize unlabeled sensor data for personalized federated learning~\cite{Bettini2021, yu2021fedhar, sarkar2021grafehty}. Meta-HAR~\cite{li2021meta} adopts Model-Agnostic Meta Learning to boost the representation ability of the shared embedding network and uses personalization for adaptation on top of it. Although these works tackle heterogeneous FL environments through personalization and representation learning, all of them share an underlying assumption that the raw data of multiple devices owned by a user can be freely exchanged among each other. In other words, these FL pipelines treat each user as a client node of federated learning, who has access to raw sensor data from all the different devices owned by them. However, this inter-device raw data communication incurs not only significant communication cost but also has privacy concerns. In contrast, \system{} does not share any raw data or features even between the devices of the same user. Instead, it leverages the time-aligned nature of data collection in the \problem{} setup to reduce the statistical heterogeneity in multi-device FL.

\subsubsection*{Applications of Federated Learning in HAR}
While many of the early use-cases of FL were related to natural language processing~\cite{47586}, visual recognition~\cite{he2021fedcv, liu2020fedvision}, and speech recognition~\cite{hard2020training} tasks, we are now also witnessing its applications in the area of human-activity recognition~\cite{Bettini2021, yu2021fedhar, sarkar2021grafehty, feng2020pmf, clusterfl, liu2021distfl}. We extend this line of work on FL in HAR, albeit in the context of multi-device environments. The evaluation of our proposed approach is done on HAR datasets containing locomotion states and activities of daily living.

\subsubsection*{Multi-Device Environments and Multi-Device HAR}
Multi-device environments offer exciting opportunities to develop accurate and generalizable sensing models by leveraging the similarities and differences across devices. This is primarily because multiple devices (e.g., a smartphone, a smartwatch, a smart glass) capture the same physical phenomenon (e.g., a user's motion activity) from different perspectives. By intelligently combining these multi-perspective datasets, we can potentially develop more robust sensory inference models. Previous research also investigates ways to improve the runtime system performance through sensor selection~\cite{kang2008seemon, kang2010orchestrator, keally2011pbn, lee2014active, zappi2008activity} and sensor fusion~\cite{ordonez2016deep, peng2018aroma, vaizman2018context, yao2017deepsense, yao2018qualitydeepsense} in multi-device environments. However, multi-device environments suffer from the well-known challenge of labeled data scarcity, in that there is often insufficient labeled sensor data for each user to be able to train a robust inference model on it. To overcome this challenge, the research community has focused on collecting multi-device datasets from multiple users and training a sensory prediction model on the aggregated dataset~\cite{sztyler2017position, Opportunity}. This methodology is called \emph{centralized training}, because the data from different users is aggregated in a central repository, and a prediction model is trained on the aggregated dataset using machine learning techniques. The obvious downside of centralized training is that it requires sharing of raw data traces from users, thereby compromising user privacy. It is also important to note that even though raw sensor data (e.g., accelerometer traces) may not seem privacy-sensitive at first, it could be used to infer various private aspects of a user's life. 
As an alternative to centralized training, we explore \textit{federated learning}~\cite{mcmahan2017communication}, a more privacy-preserving approach to collaborative training for multi-device HAR.
\section{Conclusion}
Federated learning is a crucial technique to bring intelligence to pervasive mobile and edge devices while preserving privacy of users. Many challenges yet remain in applying federated learning to real-world ubiquitous computing, e.g., high non-IID-ness, heterogeneity and scarcity of labeled training data, especially with the surge of personal data-generating devices. This paper crystallizes challenges in multi-device environments by analyzing the impact of device, user, and combined heterogeneities on a human activity recognition task. Our proposed solution, \system{},  counters statistical and system heterogeneities in MDEs, featuring accuracy- and efficiency-aware device selection strategy and model personalization. 
Evaluation in our realistic FL testbed shows improvement in inference performance over various baselines with higher $F_1$ scores and lower variance across devices, while enhancing training efficiency with reduction in invalid devices and speedup in convergence to target accuracy. Our exploration in FL for multi-device environments takes one step towards accurate, efficient, and consistent deployment of privacy-preserving machine intelligence in ubiquitous sensing applications.

\section*{Acknowledgments}
We express gratitude to the reviewers for their constructive feedback on the paper. We also thank Daniel J. Beutel and Javier Fernandez-Marques from the Flower team (\url{https://flower.dev}) for their assistance in scaling our FL experiments, and David Lindlbauer for his feedback on the paper draft. We also acknowledge the authors of ColloSSL~\cite{jain2022collossl} for sharing their dataset pre-processing code.

\bibliographystyle{ACM-Reference-Format}
\bibliography{mdfl}

\newpage
\setcounter{section}{0}
\vspace{-1cm}

\renewcommand{\thesection}{\Alph{section}}
\section{Appendix}

\subsection{Summary of Utilities}
The statistical utility can range from 0 to $\infty$, depending on the loss function, e.g., categorical crossentropy loss. Higher value means higher utility.
The system utility ranges from 0 to $log(drain^i_{th})$, and higher value means higher utility.
Thus, the time utility ranges from 0 to 1, and higher value means higher utility.

\begin{table}[h]
\small
\caption{Summary of statistical, system, and time utility used for accuracy- and efficiency-aware device selection. The upper bound of statistical utility depends on the loss function, e.g., categorical crossentropy loss}
\vspace{-0.3cm}
\label{tab:utility_info}
\begin{tabular}{@{}ccc@{}}
\toprule
\textbf{Utility} & \textbf{Range} & \textbf{Utility Interpretation} \\ \midrule
Statistical Utility & [0, $\infty$] & Higher is better\\
System Utility & [0, $log(drain^i_{th})$] & Higher is better\\
Time Utility & [0, 1] & Higher is better\\
\end{tabular}
\end{table}

\vspace{-0.5cm}
\subsection{Local Finetuning} \label{sup:local_finetuning}
\begin{figure}[h]
    \begin{minipage}{\textwidth}
        \centering
        \includegraphics[width=0.35\linewidth]{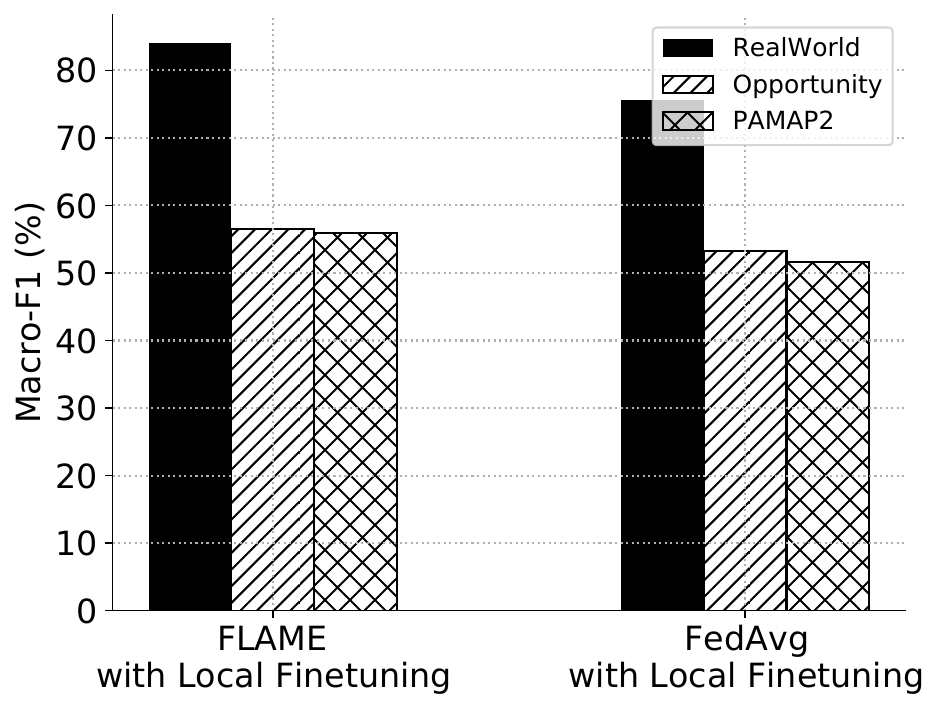}
        \vspace{-0.4cm}
        \captionof{figure}{Final macro $F_1$ scores of \system{} and FedAvg after applying local finetuning at the final round.}
        \label{fig:local_finetuning}
    \end{minipage}
\end{figure}

We ran additional experiments of adding local finetuning to \system{} and FedAvg. Local finetuning~(cf. \S3.3.3 of Kairouz et al. \cite{flsurvey}) is a technique that adds a final personalization step by training on the local dataset additionally before inference. This can be applied to FL in which a global model learns from many clients, and the model is personalized to each client at the time of usage. The results of finetuning are summarized in Figure~\ref{fig:local_finetuning}. As expected, local finetuning increases the $F_1$ scores in FedAvg---75.6\%, 53.2\%, and 51.6\% for RealWorld, Opportunity, and PAMAP2, respectively. An additional personalization achieves slightly higher or comparable the macro $F_1$ score to \system{}---84.0\%, 56.5\%, 55.9\% for RealWorld, Opportunity, and PAMAP2, respectively. However, as its device model is already personalized, the increase gap is not as large. This result reinforces the importance of personalization in FL for MDEs.

\subsection{\revision{Is the Model Learning the True Classification Problem?}}
\label{app:classification}

\parjump{}
\noindent
\revision{
In our class-based partitioning scheme, clients are inherited classes from different sources of users. This opens up a possibility for the model to learn to classify original clients instead of the true activity classes.
To clarify this concern, we investigated the features of the bottleneck layer learned by the model using TSNE plots. Our hypothesis is that if the model has inadvertently learned something about the original users, we will see clear clusters corresponding to different users in the features extracted from the data. }

\begin{figure}[h]
\centering
\subfloat[User Separability]{\label{user_tsne}\includegraphics[width=0.49\linewidth]{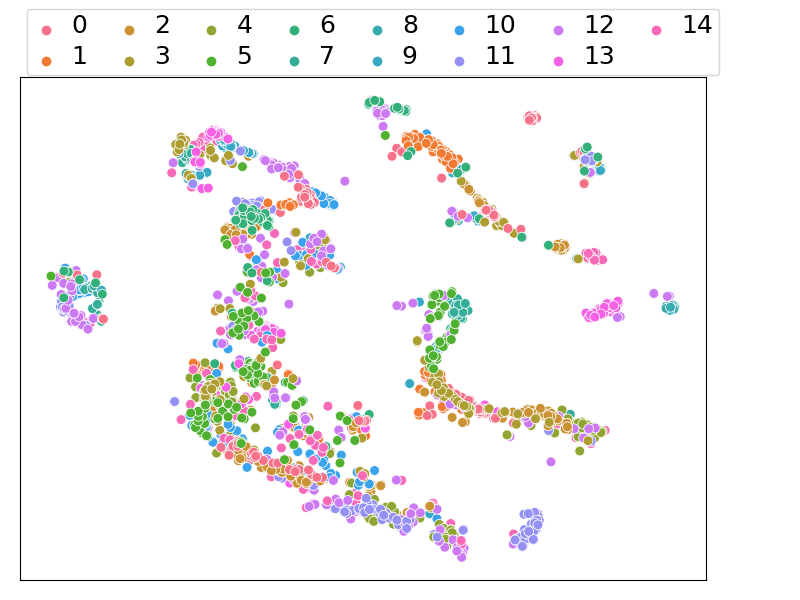}} 
\subfloat[Activity Separability]{\label{activity_tsne}\includegraphics[width=0.49\linewidth]{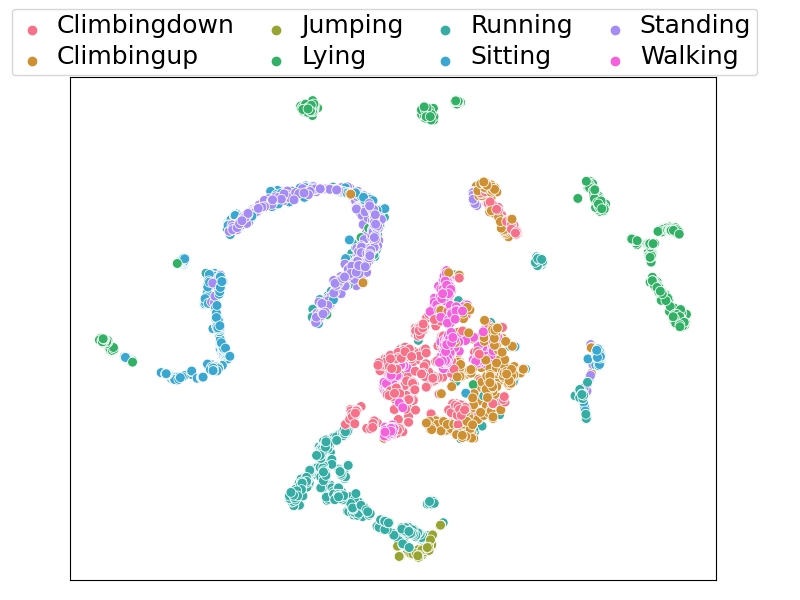}}
\caption{\revision{TSNE Plots to illustrate user and activity separability in the bottleneck features learned by the model.}}
\label{fig:tsne}
\end{figure}

\parjump{}
\noindent
\revision{Figure~\ref{user_tsne} illustrates a TSNE plot on the aggregated test set from all devices in the RealWorld dataset. Colors represent the original 15 users. We do not observe well-separated user clusters in the plot, which indicates that the model is not making its inferences based on original users. In contrast,   Figure~\ref{activity_tsne} shows a TSNE plot computed on the test data of a randomly selected participant using their personalized device model. We can clearly see clusters emerging for different activity classes, which suggests that the model has indeed learned the true activity classification problem. }

\subsection{\revision{Performance on Smartphones}}
\label{app:classification}

\revision{We have also evaluated the run-time performance of \system{} on a Google Pixel 5 device, by porting the model from Python to TensorFlow Lite and utilizing the newly released on-device training capabilities of TensorFlow Lite on Android. The results are also shown below:} 
\begin{table}[h]
\small
\caption{\revision{Training time and energy consumption per round for training an HAR model on the partitioned RealWorld dataset on a Google Pixel 5 Android device. We follow the same methodology as with other devices listed in Table~\ref{tab:energy_profiles}.}}
\vspace{-0.3cm}
\label{tab:android}
\begin{tabular}{@{}ccccc@{}}
\toprule
\textbf{Name}                                                      & \textbf{Processor Type} & \textbf{Release Year} & \textbf{\begin{tabular}[c]{@{}c@{}}Training \\ time per round \\ (in seconds)\end{tabular}} & \textbf{\begin{tabular}[c]{@{}c@{}}Energy consumption\\ per round \\ (in Joules)\end{tabular}} \\ \midrule
Google Pixel 5 & CPU & 2020 & 24.42 & 48.36\\  \bottomrule
\end{tabular}
\end{table}

\end{document}